\documentclass[journal]{IEEEtran}
\usepackage{graphicx}
\usepackage{subfig}
\usepackage{caption}
\usepackage{multirow}
\usepackage{resizegather}
\usepackage{array}
\usepackage{varwidth}
\graphicspath{ {images/} }
\usepackage{amsmath,amssymb}

\ifCLASSINFOpdf
\else
\fi
\begin{document}
%
\title{Minutia Texture Cylinder Codes for fingerprint matching}
%
%
%

\author{Wajih Ullah Baig, Umar Munir,
       Waqas Ellahi, Adeel Ejaz, Kashif Sardar
       \\(National Database And Registration Authority)
}

\maketitle

\begin{abstract}
Minutia Cylinder Codes (MCC) are minutiae based fingerprint descriptors that take into account minutiae information in a fingerprint image for fingerprint matching. In this paper, we present a modification to the underlying information of the MCC descriptor and show that using different features, the accuracy of matching is highly affected by such changes. MCC originally being a minutia only descriptor is transformed into a texture descriptor. The transformation is from minutiae angular information to orientation, frequency and energy information using Short Time Fourier Transform (STFT) analysis. The minutia cylinder codes are converted to minutiae texture cylinder codes (MTCC). Based on a fixed set of parameters, the proposed changes to MCC show improved performance on FVC 2002 and 2004 data sets and surpass the traditional MCC performance. 
\end{abstract}


%
\IEEEpeerreviewmaketitle

\section{Introduction}
%
%
%
%
\IEEEPARstart{A}{utomatic} Fingerprint Identification Systems (AFIS) are popular and concrete techniques identifications of personnel. Fingerprints are established methods for identifying and verifying individuals with hight accuracy in big scale data-bases.  There are two important parts to the process of fingerprint bio-metrics. The first part deals with feature extraction and the second part deals with matching these features. 
Broadly, feature extraction can be categorized into three distinct categories.  
1) Non-Minutiae based, 2) Minutiae based and 3) Hybrid.

Non-minutia methods \cite{NonMinutiaJain} compare fingerprints with respect
to features extracted from ridge orientation ,frequency and
texture. These use core points as the only valid entity for feature extraction. Correlation-based techniques \cite{MinutiaBasedJiang}
which compare the global pattern of ridges and furrows to see if
the ridges in two fingerprint images align, are the most prominent
example of non-minutia-based matching.
Minutia based methods try to extract minutiae and related information such as angle, position, neighborhood structure etc to produce the feature vector for the next stage of matching. Minutia Cylindrical Codes \cite{MCC} is also a minutia based technique which is very popular due to its high accuracy .
Hybrid methods \cite{HybridFM} employ both minutia and non-minutia
(e.g. ridges) features for matching. Variants of texture features
extraction per minutia can be in the form of wavelet transforms,
local binary descriptors, orientation fields, tessellations
etc. \\
Matching of features is also an active area of research. There are numerous techniques that are employed in order to improve the accuracy of the matching technique. Some techniques base their performances on statistical evidences
that were deduced during a previous step. E.g.; Kmeans-quality
clustering of minutiae \cite{kmeans-fpq} to improve
accuracy by using minutiae clusters based on minutiae quality,
\cite{biometricComb} uses the combination of different fingerprint
features to produce a more reliable matching score by introducing machine learning.\\
Feng \cite{FENG} proposed a combination of orientation and frequency features using SVM to improved matching accuracy with the overhead of training and testing SVMs. Kplet-BFS \cite{KpletBFS} matching uses a K-plet graph matching technique to match minutiae where each minutia is checked for a neighborhood of minimum k neighbors before a tree-graph match is performed.
Star \cite{starmatch} like structural matching of minutiae is used where each minutia is checked for the neighboring minutiae resulting in a star like structure.\\ 
Inspired by the concept of "relaxation" \cite{RelaxationMin} the authors of MCC incorporated relaxation as a measure of compatibility of minutiae to deduce a final penalized score. This important concept helps reducing invalid minutiae matching scores which improve the overall accuracy of  MCC.
Another minutiae based technique which shows improvements over MCC is the Complex Gaussian Mixture Model \cite{CGMM} which models minutiae in a fixed radius as mixture of Gaussian distributions.

Rest of the paper is organized as follows: Section \ref{literaturereview} is the literature review. Section \ref{motivationandcontributions} details the motivation and contribution of this work.  Section \ref{initialsteps} Explains the steps that form the basis of MTCC featurs. This includes segmentation, enhancements, orientation, frequency and energy image generation and finally minutiae extraction. Sections \ref{MCCIntro} and \ref{MTCCIntro} provide the insight into MCC and MTCC features. Matching is referred in section \ref{matching}. Results are provided in section \ref{ExperimentalResults} and conclusion is presented in \ref{conclusion}.

\section{Literature Review}
\label{literaturereview}
Minutiae based matching takes on to be at the core of most fingerprint matching techniques. Some techniques used fixed neighbor of minutiae approach in order to accurately perform a valid matching. 
The location and angle of each minutia take part in providing the discriminatory information that is comparable with other minutiae templates. This discriminatory behavior is explored by numerous researchers and have provided numerous technique of matching minutiae. Two important aspects of fingerprint matching are the feature descriptors and descriptor matching. A single descriptor can pass though variant methods of matching. Table \ref{minutiaebasedmatchingmethods} refers to a list of these descriptors.

Minutiae triplets \cite{minutiaeTriplets} takes into account the triplet pairs of minutiae to improve matching score. The technique uses a structural formation triangles from neighboring minutiae to improve the accuracy of matching.

\begin{table}[h]
\caption{Minutia based fingerprint matching variants methods}
\label{minutiaebasedmatchingmethods}
\begin{tabular}{|*{4}{V{2cm}|}}
\hline
\textbf{Method Name}     & \textbf{Minutiae Only}  &\textbf{Descriptor Length}&\textbf{Matching} \\ \hline \hline
MCC   & Yes & Fixed& Multiple matching techniques  \\ \hline
Minutiae Triplets   & Yes & Fixed& Single matching technique  \\ \hline
MRC   & Yes & Fixed& Multiple matching techniques  \\ \hline
MDC   & Yes & Fixed& Multiple matching techniques  \\ \hline
MTM   & Yes & Fixed& Multiple matching techniques  \\ \hline
Indirect local features   & Yes& Fixed & Single matching technique  \\ \hline
Point-set Registration   & Yes& Fixed & Single matching technique  \\ \hline
 \hline
\end{tabular}
\end{table}

Among minutiae based descriptors, there is a category of fixed length descriptors which are represented as vectors. These fixed length vector representations have the additional benefit of being fast and less memory intensive. Minutiae relation codes (MRC) \cite{MRC} is a technique that  represents the relationship (spatial, angular) between arbitrary minutiae againts a central minutia. Another related descriptor is the cost effective minutiae disk code \cite{MDC} which is a cost effective version of MCC. Instead using cells like MCC. the cylinder sectors are used to find "spatial and angular contributions" with reduced computational intensity as compared to MCC. A similar concept to MDC with variant matching strategies introduced by \cite{DeKock} shows how matching strategies can improve matching scores given a fixed set of descriptors. 
Minutia tensor matrix (MTM) \cite{MTM} is powerful technique that introduces tensors for  matching and retrieving minutiae pairs that are strongly related based on spectral matching of minutiae. In MTM, the diagonal elements of a similarity matrix, indicate similarities of minutia pairs where as the non-diagonal elements indicate pairwise compatibility of minutia. The non-diagonal elements in comparison to similar minutiae are checked for higher compatibility and higher scores. 
\\ Minutiae-based indirect local features \cite{FingerprintIndirectLocalFeatures} provides a 4 dimensional feature to tolerate scale, rotation and minutiae extractor errors. Using a minutiae triple feature along with order of minutiae, radial correlation and in-variance to scale as the 4D feature. 
\\In \cite{PointSetRegistration} an iterative global alignment of two minutiae sets is derived analytically using least squares method to improve matching accuracy. This is a typical case of single descriptor and the matching strategy tries to improve the accuracy of matching.  \\ 
MCC is also a fixed length vector representation of minuitae. Each minutia is encoded into a 3D data structure called cylinder. Each cylinder is further divided to cuboids or cells which hold the spatial and angular contributions of the neighboring minutiae. The radius of each cylinder being a fixed constant and thus providing the base for fixed vector length descriptor. 
There are two important parts to MCC which affects the performance of MCC based verification / identification. The first part deals with extracting the features for each minutia in a fingerprint template. 
The second part deals with matching of these cylinder. The authors of MCC provide varying matching techniques such as local simialrity sort (LSS), relaxation, normalization and localized greedy searches. These techniques are used in combination with LSS to provide high accuracy matching.
MCC is notably one of the most accurate minutiae based descriptors \cite{MCCHighTaxAcc} \cite{PerfEvalMinutiae}. 

\section{Motivation and Contributions}
\label{motivationandcontributions}

The main motivation for using texture features in MCC come from two methods related to fingerprint matching;

\begin{itemize}

  \item 
As mentioned earlier, Feng's proposed the method of using frequency and orientation as two separate descriptors of the same shape but different accuracy and combining these descriptors under the notion of SVM to improve matching accuracy.  Using the same analogy of separate descriptors of different accuracy, our method also uses the same descriptor 'shape', i.e. the minutia cylinder shape is same. The differences lie in the underlying features that are either replaced or calculated differently.
\item
STFT \cite{STFT} fingerprint enhancement is a powerful technique. The block-wise analysis of fingerprint image provides insight into three key features of each block - namely, orientation, frequency and energy. Thoroughly studied, these block-wise features have discriminative behaviour. This behaviour is empirically proved in out tests and shown in section \ref{ExperimentalResults}.

\item
Inspired by Feng's method and discriminative features from STFT, the 'shape' or the 3D data structure was a perfect fit to test out  
some modifications to MCC. These modifications include replacing minutiae angles with orientation, frequency or energy and changing the underlying equations that calculate various outputs (contributions as termed by MCC authors) for the descriptor.

\end{itemize}

The main contributions by the proposed method are as follows;

\begin{itemize}
\item 
Five variants to the MCC descriptor by incorporating texture features.  STFT analysis lays the foundation for texture features. That is, the orientation, frequency and energy features are extracted during STFT analysis. 
\item
Incorporating Successive Mean Quantize Transform (SMQT) \cite{SMQT} enhancement of fingerprint images. Light and faded ridges are made prominent using SMQT. This technique seems is a better approach than traditional techniques like simple contrast enhancements or histogram normalization. Even when viewed during visualization of SMQT results, the images produced show highly clear ridges.

\end{itemize}

\section{MCC/MTCC Basis Steps}
\label{initialsteps}

\begin{figure*}[h]
\centering
\subfloat[Input image]
{\label{inputimg}\includegraphics[height=1.3in]{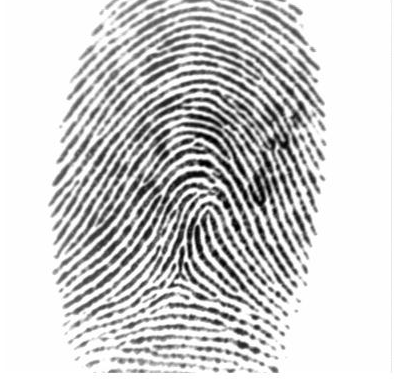}}
\thinspace\thinspace
\subfloat[Segmented mask]
{\label{mask}
\includegraphics[height=1.3in]{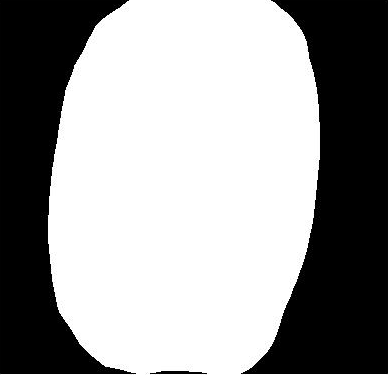}}
\thinspace\thinspace
\subfloat[STFT enhanced image.]{\label{fft_cubs_enhimg}\includegraphics[height=1.3in]{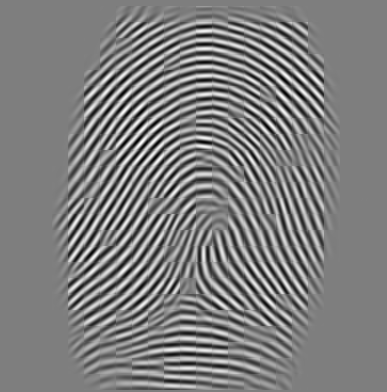}}
\thinspace\thinspace
\subfloat[Gabor enhanced image (after STFT analysis).]{\label{gabor_enhimg}\includegraphics[height=1.3in]{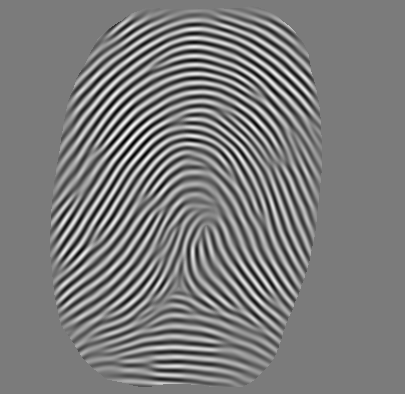}}
\thinspace\thinspace
\subfloat[SMQT normalized image. Final ouput]{\label{enhimg}\includegraphics[height=1.3in]{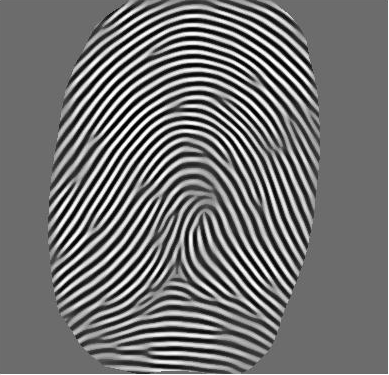}}
\thinspace\thinspace
\\
\caption{{\bf Fingerprint enhancement in proposed method.} Using prominent steps STFT, Gabor and SQMT, the proposed method produces very fine enhanced images. STFT analysis of the fingerprint image tends to make the ridges rough after joining broken ridges. The roughness is removed by Gabor filtering. SQMT helps in enhancing light ridges.}

\label{enhancementimages}
\end{figure*}

\subsection{Fingerprint image segmentation}
\label{segmentation}
The basic step of segmentation holds importance in fingerprint matching. A good extraction of the fingerprint area ensures correct minutiae extraction during the minutiae extraction phase. In this paper we use a blockwise variance based image segmenation method, \cite{FPSegmentation}. During  segmentation a mask is also created for fingerprint area. The mask is further smoothed using simple morphological operations. The mask plays an important role as the proposed method uses the mask as a constraint to mark valid fingerprint area. A similar constraint has been proposed by the authors of MCC using a convex hull. The hull is built around the boundary minutiae of a fingerprint template providing a valid area of fingerprint.  
\subsection{Fingerprint image enhancement}
Fingerprint image enhancement is an important pre-pre-processing step. The purpose of pre-processing the fingerprint image is to enhance it to a level from where minutiae can be accurately extracted. In this paper, we follow a combination of two techniques, STFT and Gabor filtering \cite{FFTGaborCombined} with the additional step of  SQMT to ensure a reliable fingerprint enhancement:
\begin{itemize}
 
  \item Short-Time-Fourier-Transform (STFT) enhancement  of the input fingerprint image. Using STFT the image is further enhanced. STFT algorithm has the capability of enhancing and joining broken ridges.
  
    \item Gabor filtering \cite{GaborFiltering} is used to smooth the ridges retrieved from STFT. STFT has the tendency to produce block discontinuities which result in images that can be jagged. This issue is resolved using Gabor filtering. Applied to STFT enhanced image, an image with more smooth ridges is produced.
    
    \item  SMQT  is applied to produce the final fingerprint image that has prominent ridges. This step is particularly useful for enhancing light ridges despite the fingerprint image being processed in prior by STFT and Gabor filtering. 
\end{itemize}

After the final step of enhancement, a normalized fingerprint image is prepared ready to be used for further feature extractions. A visualization of outputs from each basis steps is show in Figure \ref{enhancementimages}.
 
\subsection{STFT Analysis - orientation, frequency and energy image generation}
\label{STFTAnalysisFP}
\begin{figure*}[h]
\centering
\subfloat[Minutiae overlaid image]
{\label{minutiae}\includegraphics[height=1.3in]{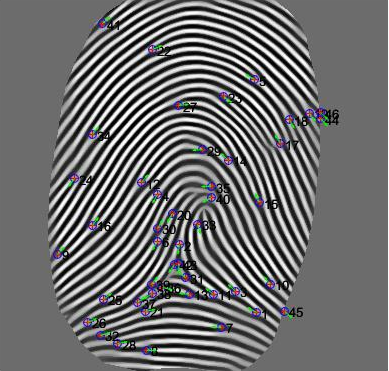}}
\thinspace\thinspace
\subfloat[STFT frequency image.]{\label{fft_cubs_fimg}\includegraphics[height=1.3in]{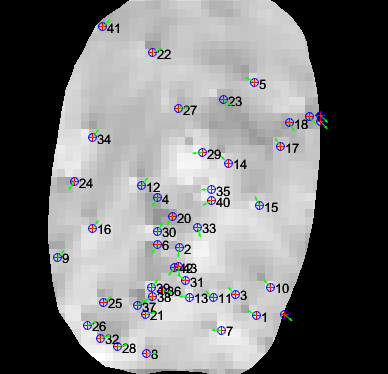}}
\thinspace\thinspace
\subfloat[STFT orientation image.]{\label{fft_cubs_oimg}\includegraphics[height=1.3in]{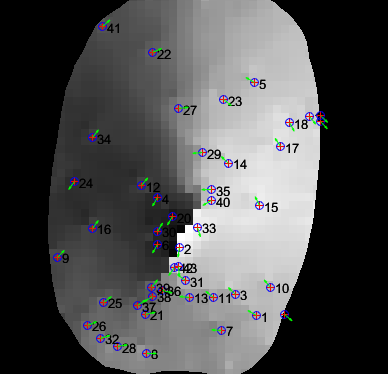}}
\thinspace\thinspace
\subfloat[STFT energy image.]{\label{fft_cubs_eimg}\includegraphics[height=1.3in]{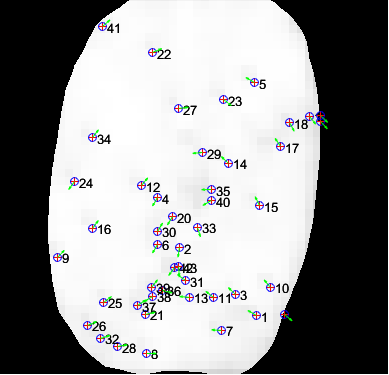}}
\thinspace\thinspace
\subfloat[Traditional orientation image.]{\label{oimg}\includegraphics[height=1.3in]{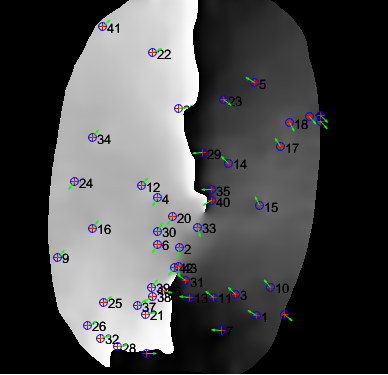}}
\\
\caption{{\bf STFT analysis images with one traditional orientation image.} STFT analysis provides useful texture information all over the image. The location of minutiae are spatial points where orientation, frequency and energy have higher differential of change which is visible in the images above. The orientation image calculated by STFT analysis differs from a traditional orientation image calculated using simple derivatives. Except for core and delta points, any local region of the fingerprint image provides consistent texture information by STFT. This is not the case in traditional gradient based orientation estimation methods.}

\label{STFTAnalysisImages}
\end{figure*}

The proposed texture features in this paper rely on STFT anaylsis of the fingerprint image. The authors of STFT consider a fingerprint image as a system of oriented texture with non-stationary properties. These non-stationary properties can be captured using contextual filtering in the Fourier domain. Therefore, to analyse the image, a 2D signal undergoes FFT analysis under a temporal window \(w(t)\). 
The authors in \cite{STFT} make use of overlapping windows of a fixed size. 
At each position of the window, a pre-defined overlap step is associated with the window \(W(x,y)\). This overlapping ensures ridge connectivity and removes 'block' effects. This also helps in more localized contextual processing of information as the fingerprint image has a varying collection of information.Each such analysis frame yields a single value which the dominant frequency, orientation or energy. In the proposed method, this single value information is replicated back onto the processing window to reproduce orientation, frequency and energy images. Thus STFT analysis yields three distinct images, namely orientation (\(I_o\)), frequency (\(I_f\)) and energy (\(I_e\)) images. Note that the energy is actually the logarithmic value of the single energy value yielded in the temporal window. Figure \ref{STFTAnalysisImages} show outputs from STFT and traditional orientation calculation.

\subsection{Minutiae extraction}
\label{miniutiaextraction}
FingerJetFXOSE \cite{FINGERJETFX} is a open source cross platform sdk that has been used to extract minutiae from the fingerprint enhancement steps. FingerJetFXOSE uses block-wise fft image enhancement which is turned off in our tests as it is not required. The minutiae list extracted by FingerJetFXOSE are sorted based on their quality.
\section{Minutia Cylindrical Codes}
\label{MCCIntro}
\begin{figure}[h]
\centering
\subfloat[A 3D cylinder]{\label{3dcylinder}\includegraphics[height=1.7in]{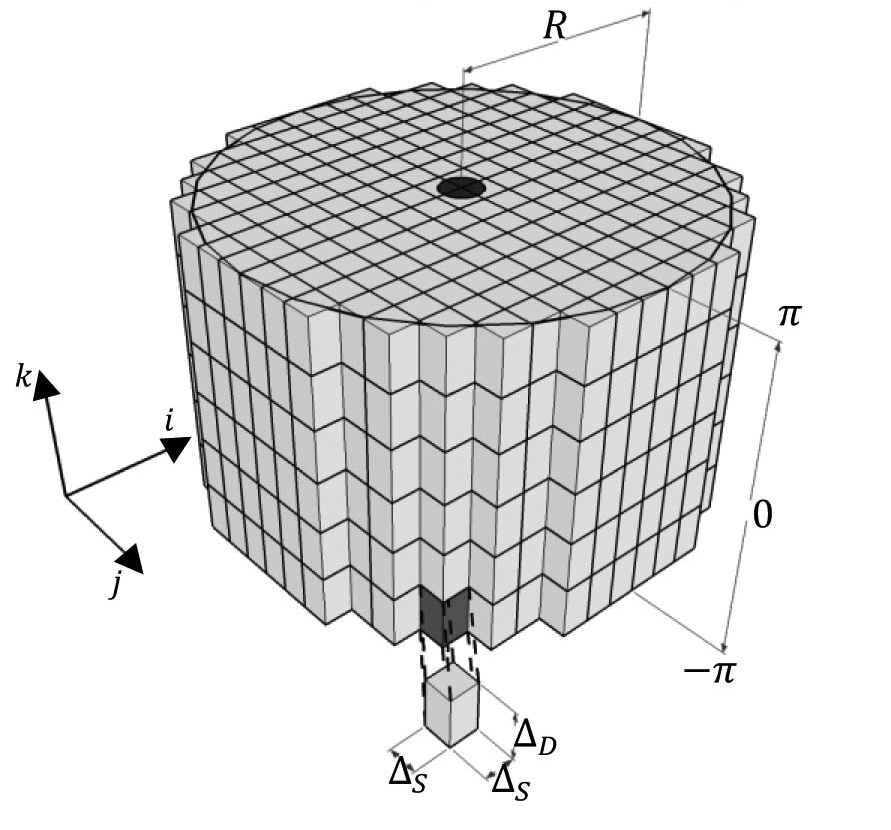}}
\thinspace\thinspace
\subfloat[A cylinder slice with cell and its neighbouring minutiae]{\label{SliceCellNeighbours}\includegraphics[height=1.7in]{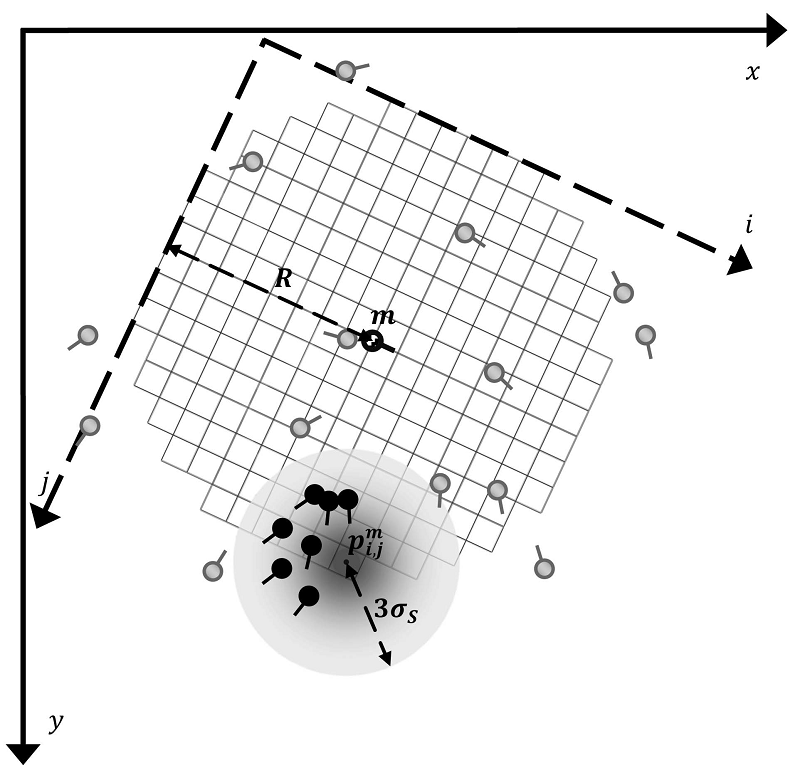}}
\\
\subfloat[A Cylinder split into sections.]{\label{cylinderSplit}\includegraphics[height=1.7in]{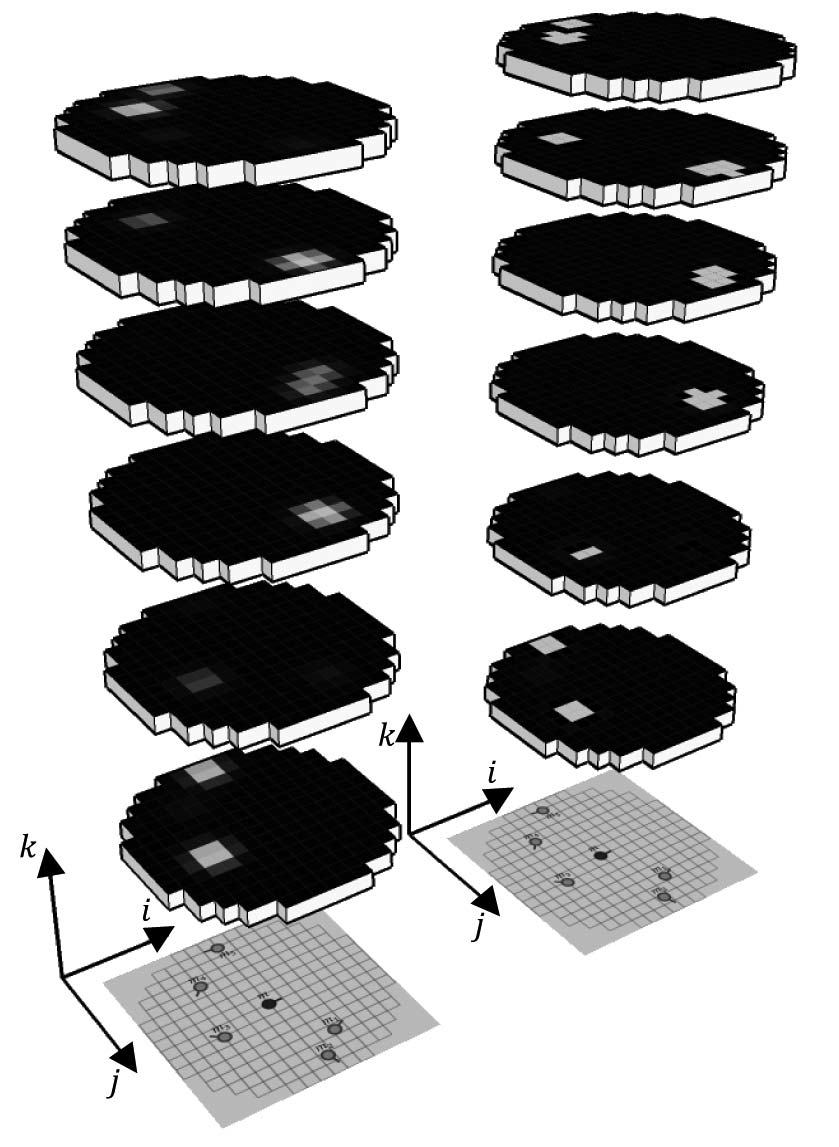}}
\\
\caption{{\bf The 3D data structure.} A 3D cylinder which forms the MCC descriptor for a a single minutia. The local structure associated to a given minutia  \(m=\{x_m,y_m,\theta_m\}\)
is represented by a cylinder with radius R and
height 2\(\pi\) whose base is centered on the minutia location
\((x_m,y_m)\) (a). Each section contains cells that hold contributions of neighboring minutiae, (b). The cells with brighter parts have more contributions due to strong neighborhoods,(c).}

\label{MCCCylinders}
\end{figure}

In this section, a brief explanation of MCC is provided with an explanation of the equations that govern MCC. Some of these equations are modified to provide the underlying changes that form the contents of MTCC descriptor.\\ MCC uses 3D data structures called cylinders (see Figure \ref{MCCCylinders}). Each cylinder is oriented in the direction of the central minutia over which it is created. We call this the default feature and denote it with \(MCC_{o}\). 

For a fingerprint template of \(n\) minutiae, \(T = \{m_1,m_2,m_3,...,m_n\}\) where as each \(m\) represents a single minutia in the form \((x_m,y_m,\theta_m)\). 
Figure \ref{3dcylinder} represents a cuboid. The cuboid is divided into \(N_c\) distinct cells where \(N_C = N_S \times N_S \times N_D \). \(N_S\) and \(N_D\) are predefined fixed constants. 
An individual cell itself is a mini-cuboid with \(\Delta_S \times \Delta_S\) as its base and \(\Delta_D\) as its height. Where as \(\Delta_S = 2R/N_S\) and \(\Delta_D = 2\pi / N_D\)
\(R\) is the fixed radius from the central minutiae around which the cylinder is created. The general concept behind a cell is the accumulation of spatial and directional contributions of the minutiae that lie around the central minutia.

Individual cells are identifiable by indices \(i,j,k\) in the cuboid. First, each cell has an associated angle to introduced by the following equation;

\begin{equation}\label{CellAngle} 
d_{\phi k}= -\pi + (k-\frac{1}{2}) . \Delta_D
\end{equation}  

Second, the center of each cell \(p^m_{i,j}\) from the location (\(x_m, y_m\)) of minutia \(m\) is calculated  as follows;

\begin{equation}\label{CellCenter} 
p^m_{i,j} = \begin{bmatrix}
      x_m \\[0.3em]
        y_m \\[0.3em]
     \end{bmatrix}
     + \Delta_S . \begin{bmatrix}
      cos(\theta_m) & sin(\theta_m) \\[0.3em]
      -sin(\theta_m) & cos(\theta_m)
     \end{bmatrix} . \begin{bmatrix}
      i - \frac{N_S+1}{2} \\[0.3em]
      j - \frac{N_S+1}{2} 
     \end{bmatrix}
\end{equation}  

For a cell at (i,j,k), a numerical value \(C_m(i,j,k)\) is calculated. This is what is called a contribution.

\begin{equation}\label{CellContribution} 
C_m(i,j,k)= \Psi \Bigg(\sum_{m_t \in N_{p^m_{i,j}}}(C_{m}^{S}(m_t,p^m_{i,j}))*(C_{m}^{D}(m_t,d_\phi k))\Bigg)  
\end{equation}

\begin{equation}\label{PSI} 
 \Psi = Z(u,\mu_\Psi\ ,\tau_\Psi)
\end{equation}

In equation \ref{CellContribution}, \(C_{m}^{S}\) and \(C_{m}^{S}\) are the spatial and directional contributions of neighboring minutiae \(N_{p^m_{i,j}}\) that are in a radius \(3\sigma_S\) from the center of the cell \(p^m_{i,j}\). The contribution is controlled by a sigmoid function that ensures the contributions are between 0 and 1.

\begin{equation}\label{CellNeighbourhood} 
N_{p^m_{i,j}} = \{m_t \in T; m_t \neq m, d_S(m_t,p^m_{i,j}) \leq 3 \sigma s \}
\end{equation}

The contribution of each minutiae in \(N_{p^m_{i,j}}\) is given as the product of the spatial and directional contributions.

\begin{equation}\label{SpatialContribution} 
C_{m}^{S}(m_t,p^m_{i,j}) = G_S(d_S(m_t,p^m_{i,j}))
\end{equation}  

\begin{equation}\label{DirectionalContribution} 
C_{m}^{D}(m_t,d_\phi k) = G_D(d_\emptyset(d_{\phi k},d_{\theta}(m,m_t)))
\end{equation}  
		
\(G_S\) and  \(G_D\) are the Gaussian functions in spatial and directional contributions domain.  In equation \ref{DirectionalContribution} \(d_{\theta}(m_1,m_2)\) is the normalized difference of angles between two minutiae and is defined as follows;
\begin{equation}\label{AngleNormalized1} 
d_\emptyset(\theta_1,\theta_2) = \begin{cases}
    \theta_1 - \theta_2  & if   -\pi  \leqslant \theta_1 - \theta_2 < \pi\\
   2\pi + \theta_1 - \theta_2  & if   \theta_1 - \theta_2 < -\pi\\   
   -2\pi + \theta_1 - \theta_2  & if   \theta_1 - \theta_2 \geqslant \pi\\
  \end{cases}
\end{equation}  

\begin{equation}\label{AngleNormalized2} 
d_\theta(m_1,m_2) = d_\emptyset(\theta_{m_1},\theta_{m_2})
\end{equation}  

Not every cell in a cylinder accumulates contribution according to equation \ref{CellContribution}. Cells that lie outside the valid mask area are considered to be invalid and cells with no neighbors have zero contribution. 

A cylinder is kept or discarded under the constraints of validity. These constraints include a minimum number of neighbors around the central minutia under a constant radius,percentage count of the total number of valid cells \cite{MCC}. Only valid cylinders are the part of the fingerprint template.

\section{Minutia Texture Cylinder Codes}
\label{MTCCIntro}
Taking into account the description of MCC in the previous section, the proposed features are presented in the two sets. Texture feature sets 1 and 2. Set 1 being local features and set 2 being global features. Note that the terms local and global are with respect to the MCC descriptor.
\subsection{Texture features set 1, Local Frequency and Energy features}
The first set of auxiliary features replacing \(MCC_{o}\) are 
\(MCC_{f}\) and \(MCC_{e}\) which stand for MCC frequency and MCC energy features.
\begin{figure}[h]
\centering
\subfloat[ A single slice of a cylinder for minutia \#1 of image 1\_1.tif (enhanced) from FVC 2002 DB1A. Only part of the image is show for brevity.]
{\label{cylinderSlice}\includegraphics[scale=0.7]{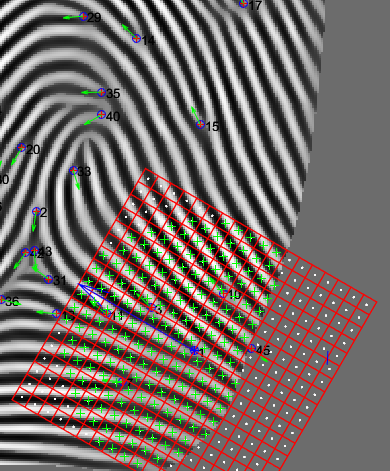}}
\\
\caption{{\bf A cylinder slice shown over a minutia} The cylinder slice is made up of cells that marked with  '+' for valid and a  '.' for invalid. Cells outside the image area are truncated for visualization. These are also marked invalid. The blue arrow shows the direction of the central minutia over which the cylinder is created.}

\label{CylinderSingleSlice}
\end{figure}

The replacement of the default features is pretty simple. The notion is to replace the angular contributions with frequency or energy contributions. 

Considering equation \ref{DirectionalContribution}, only the directional contribution part is modified to deduce the first independent texture feature based on frequency. These are the proposed \(MCC_{f}\) features.

\begin{equation}\label{FrequencyContribution} 
C_{m}^{F}(m_t,d_\phi k) = G_F(d_\emptyset(d_{\phi k},d_{\theta}(I_f(x_m,y_m),I_f(x_{m_t},y_{m_t}))))
\end{equation}

Equation \ref{FrequencyContribution} is similar to \ref{DirectionalContribution} except that instead of the minutiae angles for \(m\) and \(m_t\) , the frequencies at the locations \(I_f(x_m,y_m)\) and \(I_f(x_{m_t},y_{m_t})\) are picked up from \(I_f\) image respectively. Hence, for each cell a spatial and a frequency contribution is calculated. Since the frequency image has already been normalized between \(-\pi\) and \(\pi\), this perfectly sets the data to be used by equation \ref{AngleNormalized1}. 

Similar to \(MCC_{f}\) features, \(MCC_{e}\) features are introduced. 
Again, considering equation \ref{DirectionalContribution}, only the directional contribution is modified.

\begin{equation}\label{EnergyContribution} 
C_{m}^{E}(m_t,d_\phi k) = G_E(d_\emptyset(d_{\phi k},d_{\theta}(I_e(x_m,y_m),I_e(x_{m_t},y_{m_t}))))
\end{equation}  

\begin{figure*}[h]
\centering
\subfloat[\(MCC_o\) Default  cylinder slice.]
{\label{mcc_o}\includegraphics[height=1.5in]{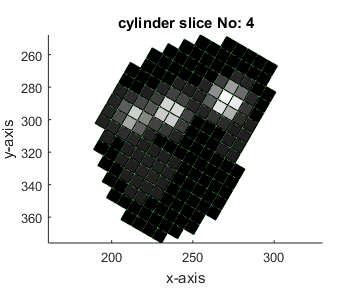}}
\thinspace\thinspace
\subfloat[\(MCC_f\) frequency cylinder slice.]{\label{mcc_f}\includegraphics[height=1.5in]{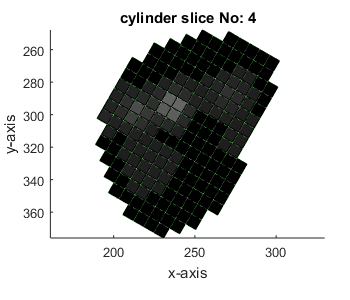}}
\thinspace\thinspace
\subfloat[\(MCC_e\) energy cylinder slice.]{\label{mcc_e}\includegraphics[height=1.5in]{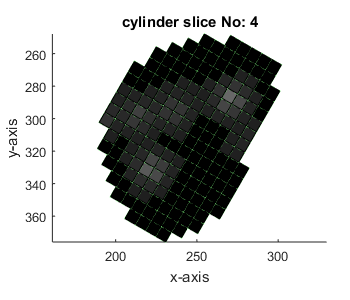}}
\\
\caption{{\bf Minutia cylindrical codes. A single section from the cylinder are shown for each feature.} The cylinder slices shown here represent contributions from minutia angle (default feature), minutiae local frequency (texture feature) and minutiae local energy (texture feature). The contributions vary for each feature.}

\label{MCCLocalTexturesFeatures}
\end{figure*}

\subsection{Texture features set 2, Cell Centered Orientation, Frequency and Energy features}
The previous section dealt with texture features that were mere replacements of the minutiae directional contributions. In this section, a major change in the way the contributions are calculated is proposed. These features are termed as cell centered orientation  \(MCC_{co}\), frequency \(MCC_{cf}\) and energy \(MCC_{ce}\) features. 

Spatial contributions are calculated exactly the same as described in previous section. Unlike calculating the angular contributions occurring due to the neighbors \(N_{p^m_{i,j}}\) of \(p^m_{i,j}\), the contributions are only calculated due to the center of the cell, i.e. \(p^m_{i,j}\). This means that for orientation, frequency and energy features, the center of the cell serves as a point from where the orientation, frequency or energy are picked up. Intuitively this means using \(I_o\), \(I_f\) and \(I_e\) respectively.

From equation \ref{CellContribution}, the directional contribution part is modified to include the orientations at the center of the cell \(p^m_{i,j}\) instead of the minutiae angles of the neighborhood \(N_{p^m_{i,j}}\).

\begin{equation}\label{CellWiseOrientationContribution} 
C_{m}^{D}(m_t,d_\phi k) = G_D(d_\emptyset(d_{\phi k},d_{\theta}(m,I_o(p^m_{i,j}))))
\end{equation}  

Likewise for  frequency and energy features we have the following equations;

\begin{equation}\label{CellWiseFrequencyContribution} 
C_{m}^{F}(m_t,d_\phi k) = G_F(d_\emptyset(d_{\phi k},d_{\theta}(I_f(x_m,y_m),I_f(p^m_{i,j}))))
\end{equation}

\begin{equation}\label{CellWiseEnergyContribution} 
C_{m}^{E}(m_t,d_\phi k) = G_E(d_\emptyset(d_{\phi k},d_{\theta}(I_e(x_m,y_m),I_e(p^m_{i,j}))))
\end{equation}

\begin{figure*}[h]
\centering
\subfloat[\(MCC_{co}\) cell centered orientation cylinder slice.]
{\label{mcc_co}\includegraphics[height=1.5in]{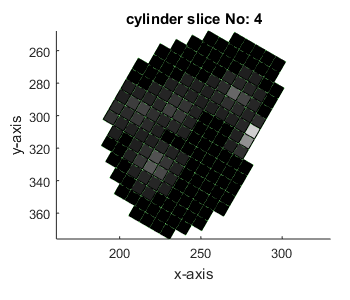}}
\thinspace\thinspace
\subfloat[\(MCC_{cf}\) cell centered frequency cylinder slice.]{\label{mcc_cf}\includegraphics[height=1.5in]{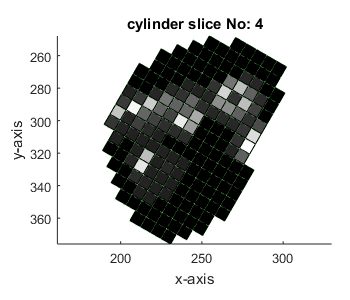}}
\thinspace\thinspace
\subfloat[\(MCC_{ce}\) cell centered energy cylinder slice.]{\label{mcc_ce}\includegraphics[height=1.5in]{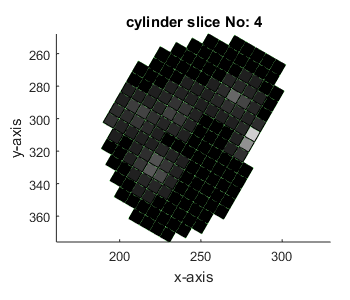}}
\\
\caption{{\bf Minutia cylindrical codes. A slice from the cylinder is shown for each feature} The cylinder slices shown here represent contributions for cell centered orientation, frequency and energy features (all being texture features). Evident that the contributions vary for each feature.}

\label{MCCCellCenterdTexturesFeatures}
\end{figure*}

The single slice shown in figure \ref{CylinderSingleSlice} is part of a full cylinder. The same slice is used to calculated the MTCC features shown figures  \ref{MCCLocalTexturesFeatures} and  \ref{MCCCellCenterdTexturesFeatures}.

\section{Matching - Local Similarity Sort with Relaxation.}
\label{matching}

The authors of MCC incorporate "relaxation" of matched minutiae to deduce compatible pairs. This compatibility is actually a penalization method for structurally dissimilar minutiae pairs. The more dissimilar a pair, the higher the penalization. The method is abbreviated as "LSSR". That is, Local Similarity Sort with Relaxation. \\
A local similarity matrix LSM of size \(nA \times nB\) is produced by matching all the reference minutiae \(nA\) with all the minutiae \(nB\) of the query template. LSM is sorted to pick top \(nr\) matching minutiae that are passed onto relaxation. Once penalized by relaxation, top \(np\) from top \(nr\) minutiae are selected to produce the final matching score.

\section{Experimental Results}
\label{ExperimentalResults}
In order to evaluate the MTCC features,  we use databases DB1A, DB2A, DB3A and DB4A of FVC2002 and FVC2004 competitions. These databases are commonly used as benchmarks for evaluating fingerprint matchers in the context of fingerprint verification. For all the databases, each fingerprint impression is matched against the other impression of the same finger to compute false rejection rate (FRR) and the false acceptance rate (FAR). The total number of genuine tests for DB1A will be [(8x7)/2] x 100 = 2800. On the other hand, the first impression of each finger is matched against the first impression of other fingers of the same database to computer false acceptance rate (FAR). Thus for DB1A we will have of [(100x99)/2] x 100 = 4950 impostor scores. It should be noted that the image quality in the four parts/datasets is different as they were acquired using different devices. (See Tables \ref{FVC2002capturingInfo} and \ref{FVC2004capturingInfo})

\begin{table}[h]
\caption{FVC 2002 DB parts - imaging details}
\label{FVC2002capturingInfo}
\begin{tabular}{|*{4}{V{2.8cm}|}}
\hline
\textbf{DB}     & \textbf{Device} & \textbf{Dimensions }& \textbf{Resolution (DPI)} \\ \hline \hline
1A   & Optical/TouchView II - Identix & 388 x 374 & 500  \\ \hline
2A   & Optical/FX2000 - Digital Persona & 300 x 300 & 569   \\ \hline
3A 	& Capacitive Sensor/100 SC - Precise Biometrics & 300 x 300 & 512    \\ \hline
4A   & Synthetic - SFinGe v2.5 & 288 x 384 & 500 \\ \hline
\end{tabular}
\end{table}

\begin{table}[h]
\caption{FVC 2004 DB parts - imaging details}
\label{FVC2004capturingInfo}
\begin{tabular}{|*{4}{V{2.8cm}|}}
\hline
\textbf{DB}     & \textbf{Device} & \textbf{Dimensions }& \textbf{Resolution (DPI)} \\ \hline \hline
1A   & Optical/V300 - Crossmatch & 640 x 480 & 500  \\ \hline
2A   & Optical/U.are.U 4000 - Digital Persona & 328 x 364 & 500   \\ \hline
3A 	& Capacitive Sensor - FingerChip FCD4B14CB & 300 x 480 & 512    \\ \hline
4A   & Synthetic - SFinGe v3.0 & 288 x 384 & 500 \\ \hline
\end{tabular}
\end{table}

\subsection{Experimental Settings}
We carry out all the experiments on a desktop machine with an Intel Core i7 4790 processor (3.60 GHz) and 16 GB of RAM. Our experiments are rely on  a Matlab code-base for the proposed MTCC features. Therefore we do not perform a speed comparison with .NET code (MCC SDK).

The following settings are in place for all tests; 

\begin{itemize}

\item Fingerprint image enhancement are applied to all databases. The choice of the temporal window  is at 14 \(\times\) 14 pixels with an overlap of 6 pixels in STFT analysis of the fingerprint images.

\item Distance metrics. Another important factor that lies at the core of accuracy is the euclidean distance. We also provide sine and cosine replacements for euclidean distance in texture feature matching. The authors of MCC proposed two variants of the euclidean distance to match two cylinders. A floating point variant and a bit-based variant. Floating point being more accurate while bit-based most useful for fast and low-end embedded systems. 
In addition to euclidean distance which is used only for matching of \(MCC_o\) features, the double angle sine and cosine floating point distance metrics are used to match rest of the features, i.e. \(MCC_f\),\(MCC_e\),\(MCC_{co}\),\(MCC_{cf}\),\(MCC_{ce}\). Double angles are used because \(\theta  + \pi\) and \(\theta\) have exactly the same meaning in orientation representation  or is a mapping of the double angle representation maps an angle \(\theta\) and its mirror angle \(\theta  + \pi\) to the same angle \cite{VisionWithDirection}.

\begin{equation}\label{EuclideanDistance} 
d_{(C_a,C_b)} = \begin{cases}
    1-\frac{||C_{a|b}-C_{b|a}||}{||C_{a|b}||+||C_{b|a}||}  & if\ matchable\\
  0  & otherwise\\  
  \end{cases}
\end{equation}  

\begin{equation}\label{DoubleCosAngleDistances} 
Cos_{d_{(C_a,C_b)}} = \begin{cases}
    1-\frac{||cos(2C_{a|b})-cos(2C_{b|a})||}{||cos(2C_{a|b})||+||cos(2C_{b|a})||}  & if\ matchable\\
  0  & otherwise\\
  \end{cases}
\end{equation}  

\begin{equation}\label{DoubleSinAngleDistances} 
Sin_{d_{(C_a,C_b)}} = \begin{cases}
    1-\frac{||sin(2C_{a|b})-sin(2C_{b|a})||}{||sin(2C_{a|b})||+||sin(2C_{b|a})||}  & if\ matchable\\
  0  & otherwise\\
  \end{cases}
\end{equation}  
\begin{equation}\label{DoubleAngleDistance} 
d_{\gamma({C_a|C_b})} = \frac{\sqrt{Cos_{d_{(C_a,C_b)}}^2+Sin_{d_{(C_a,C_b)}}^2}}{2}
\end{equation}  
 Matching of two cylinders is undertaken by valid and aligned cells. Invalid cells in both cylinders are discarded and only valid cells take part in calculation of distance. A Euclidean distance (equation \ref{EuclideanDistance}) for \(MCC_O\) features and double angle distances (equation \ref{DoubleAngleDistance}) for texture features. The matching of two cylinder is constrained by a global rotation angle \(\delta_\Theta\) and a minimum number of valid match-able cells \(min_{ME}\).

\item LSS-R matching selected for all tests.

\item Fixed set of parameters for all tests. These are defined in Table \ref{ConstantsTable}
\begin{table}[!ht]
\centering
\caption{
{\bf Parameters for MCC/MTCC features.}}
\begin{tabular}{|*{2}{p{2cm}|}}
\hline

\textbf{Constant} & \textbf{Value} \\ \hline \hline
 R & 65 \\ \hline
 \(N_S\) & 18  \\ \hline
 \(N_D\) & 5 \\ \hline
 \(\sigma_S\) & 6 \\ \hline
\(\sigma_D\) & \(\frac{5}{36}\pi\)\\ \hline
\(\Omega\)\(^1\)& 0\\ \hline
\(\mu_\Psi\) & \(\frac{5}{1000}\)\\ \hline
\(\tau_\Psi\) & 400\\ \hline
\(min_{vc}\) & 0.20\\ \hline
\(min_{M}\) & 1\\ \hline
\(min_{ME}\) & 0.20\\ \hline
\(\delta_\theta\) & \(\frac{2}{3}\pi\)\\ \hline
\(\mu_p\) & 30\\ \hline
\(\tau_p\) & \(\frac{2}{5}\)\\ \hline
\(min_{NP}\) & 4\\ \hline
\(max_{NP}\) & 10\\ \hline
\(w_R\) & 0.6\\ \hline
\(\mu_{1}^{p}\) & 12\\ \hline
\(\tau_{1}^{p}\) & -0.8\\ \hline
\(\mu_{2}^{p}\) & \(\frac{\pi}{12}\)\\ \hline
\(\tau_{2}^{p}\) & -30\\ \hline
\(\mu_{3}^{p}\) & \(\frac{\pi}{28}\)\\ \hline
\(\tau_{3}^{p}\) & -10\\ \hline
\(N_{rel}\) & 4\\ \hline
\end{tabular}
\begin{flushleft} Parameters used in performing tests across FVC DBs. \\
\(^1\)  \(\Omega\) parameter is used to expand the convex hull formed by the out-lying minutiae in fingerprint template. This parameter is set to zero in our tests which mimics the same affect of fingerprint mask used in this paper.
\end{flushleft}
\label{ConstantsTable}
\end{table}
 \end{itemize}

\subsection{Evaluations}
Empirical evaluations of all the features were carried for FVC 2002/2004 data-sets for the following;

\begin{itemize}

\item MCC-SDK (.NET) 
\item Our implementation of MCC features (MATLAB)
\item MTCC features (MATLAB)
\end{itemize}

\subsection{Performance Comparison}
\label{performancecomparison}

\begin{figure*}[h]
\centering
\subfloat[DET curves for FVC 2002 DB1A]
{\label{det2002db1a}\includegraphics[height=2.5in]{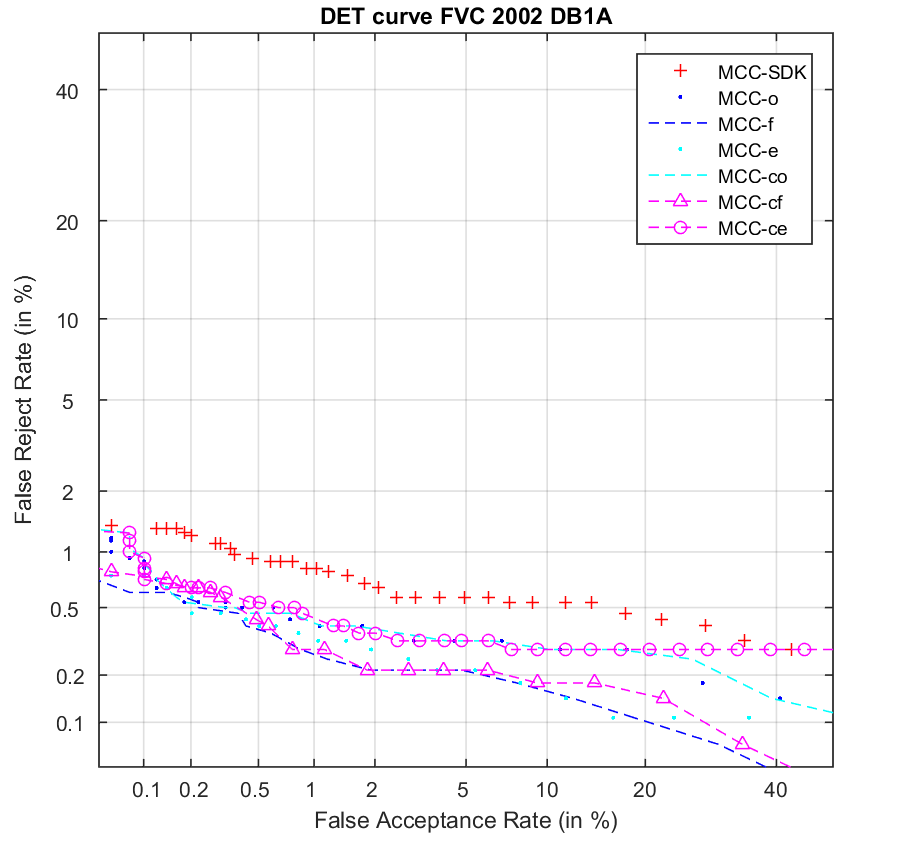}}\thinspace
\subfloat[DET curves for FVC 2002 DB2A ]{\label{det2002db2a}\includegraphics[height=2.5in]{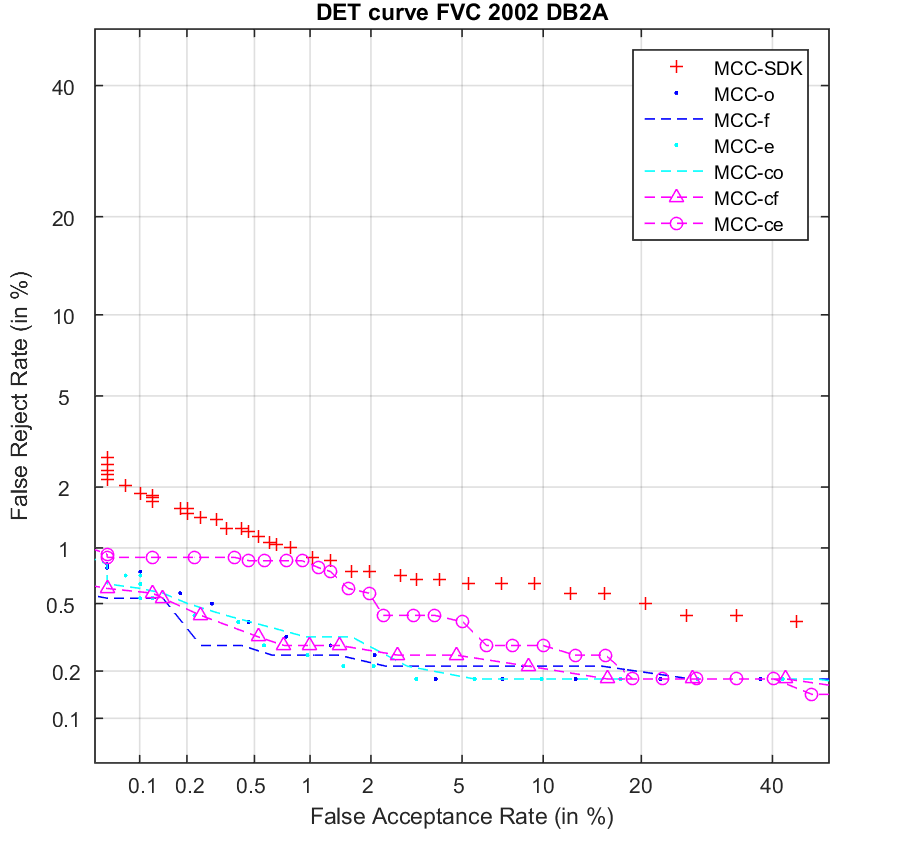}}
\\
\subfloat[DET curves for FVC 2002 DB3A]
{\label{det2002db3a}\includegraphics[height=2.5in]{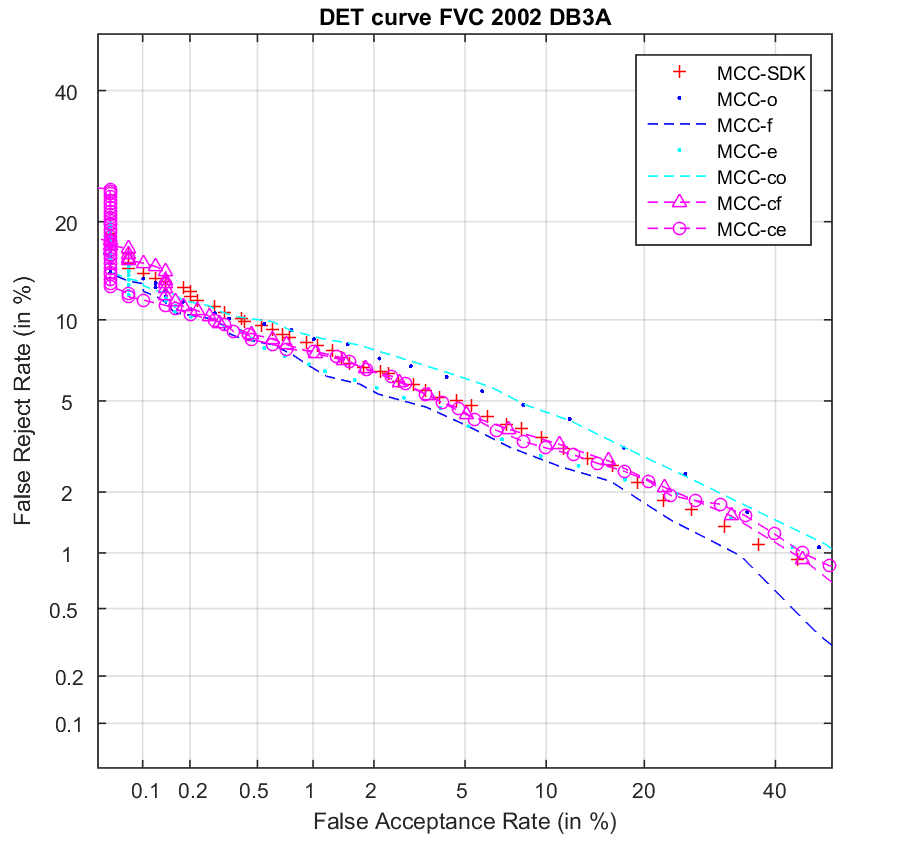}}\thinspace
\subfloat[DET curves for FVC 2002 DB4A]{\label{det2002db4a}\includegraphics[height=2.5in]{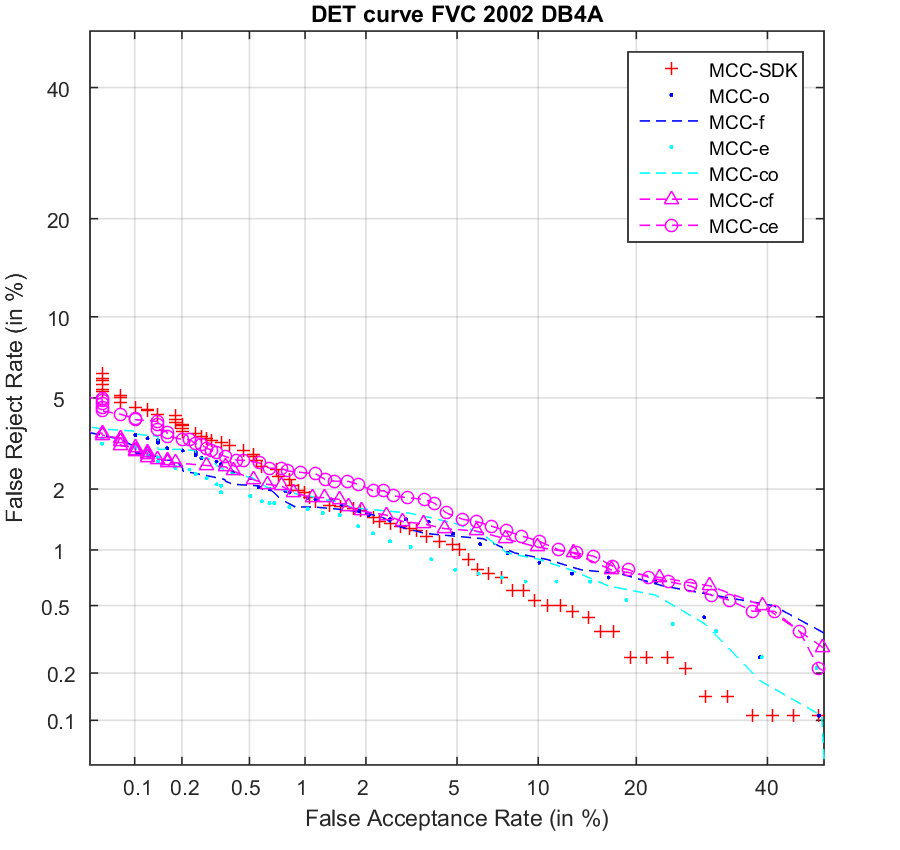}}
\\
\caption{{\bf DET curves for 2002 DBs} DET curves for all features.}

\label{DETCurves2002}
\end{figure*}

\begin{figure*}[h]
\centering
\subfloat[DET curves for FVC 2004 DB1A ]
{\label{det2004db1a}\includegraphics[height=2.5in]{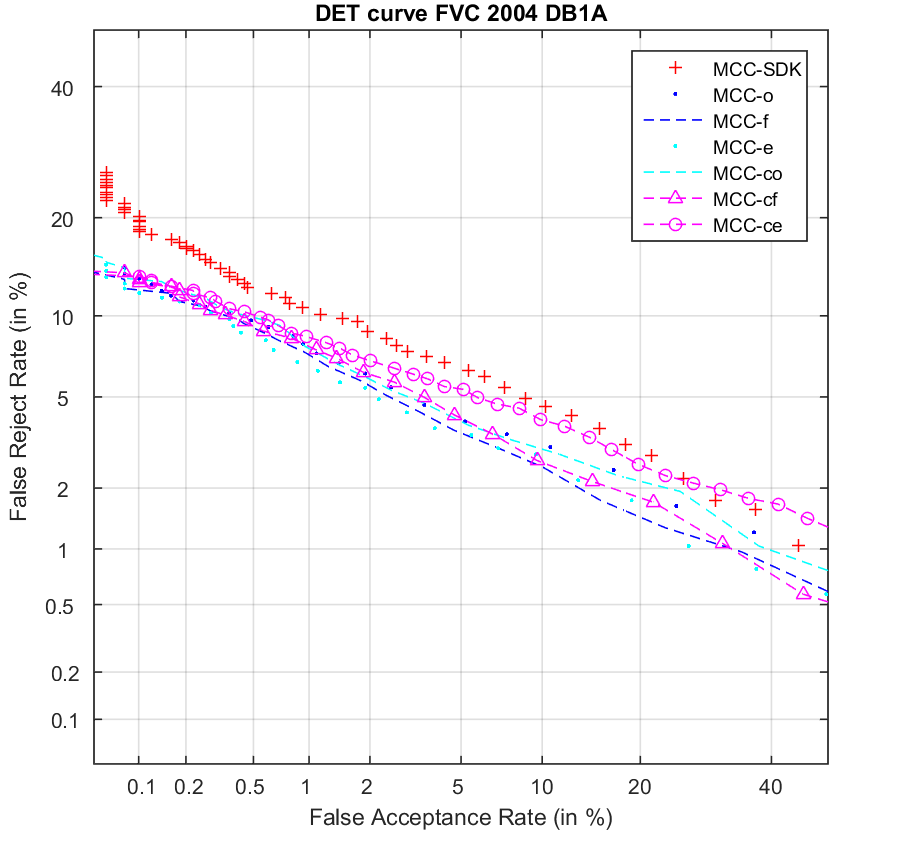}}\thinspace
\subfloat[DET curves for FVC 2004 DB2A ]{\label{det2004db2a}\includegraphics[height=2.5in]{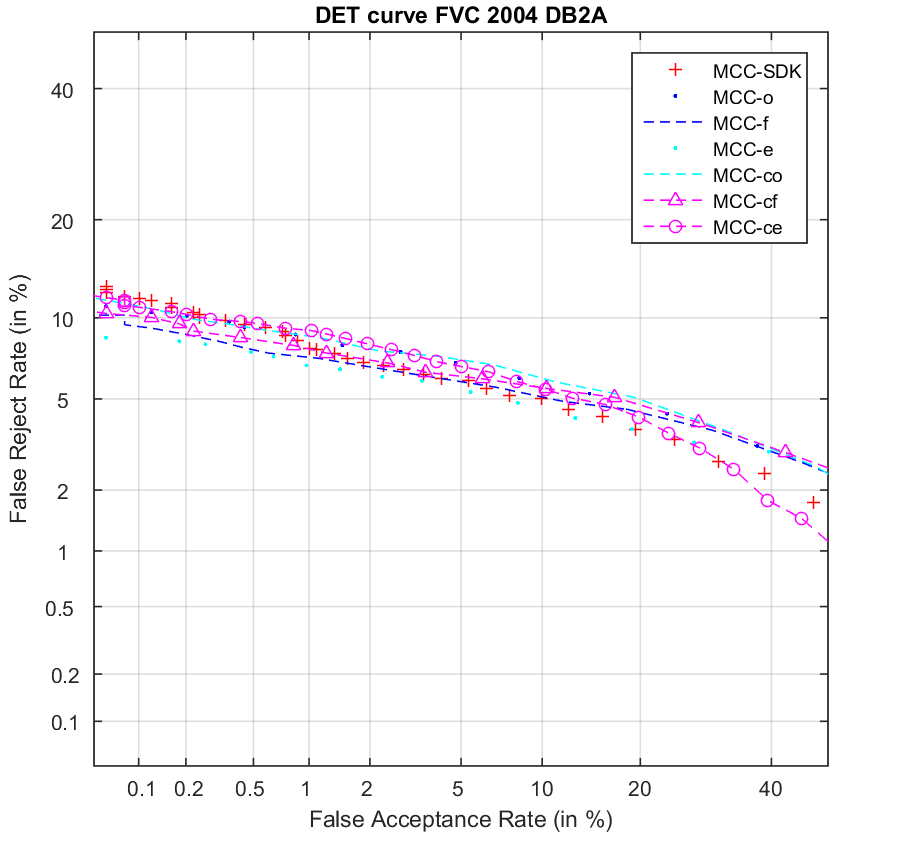}}
\\
\subfloat[DET curves for FVC 2004 DB3A ]
{\label{det2004db3a}\includegraphics[height=2.5in]{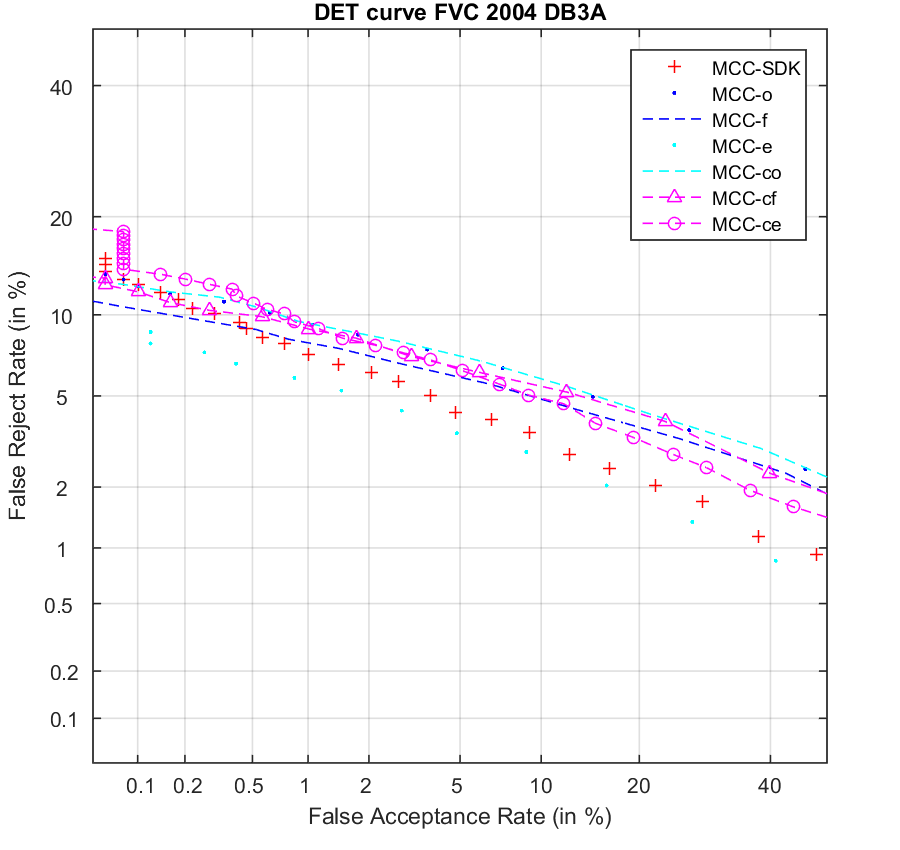}}\thinspace
\subfloat[DET curves for FVC 2004 DB4A ]{\label{det2004db4a}\includegraphics[height=2.5in]{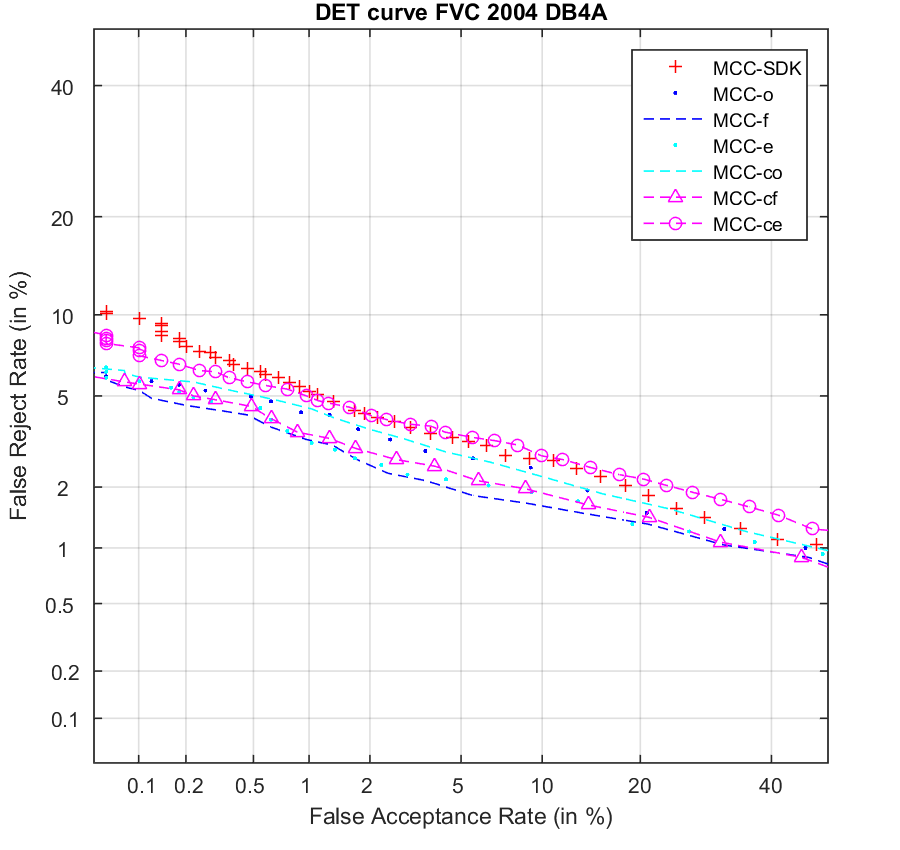}}
\\
\caption{{\bf DET curves for 2004 DBs} DET curves for all features.}

\label{DETCurves2004}
\end{figure*}

\begin{table}[h]
\caption{EERs  for FVC 2002 DBs.}
\label{eersFVC2002}
\begin{tabular}{|*{5}{p{1.11cm}|}}
\hline
\textbf{Feature}     & \textbf{DB1A} & \textbf{DB2A }& \textbf{DB3A} &\textbf{ DB4A} \\ \hline \hline
\(MCC_{SDK}\)   & 0.89 & 0.9 & 4.92 & 1.61  \\ \hline
\(MCC_o\)   & 0.54 & 0.85 & 4.68 & 2.03  \\ \hline
\(MCC_f\) 	& 0.46 & 0.36 & 5.89 & 1.63   \\ \hline
\(MCC_e\)   & 0.46 & 0.38 & 4.67 & 1.71  \\ \hline
\(MCC_{co}\)  & \textbf{0.42} & 0.38 & \textbf{4.42} & \textbf{1.49}  \\ \hline
\(MCC_{cf}\) 	& \textbf{0.42} & \textbf{0.28 }& 4.24 & 1.61   \\ \hline
\(MCC_{ce}\)  & 0.5 & 0.39 & 5.6 & 1.67  \\ \hline
\end{tabular}
\end{table}

\begin{table}[h]
\caption{FMR1000 for FVC 2002 DBs.}
\label{1000FMRsFVC2002}
\begin{tabular}{|*{5}{p{1.11cm}|}}
\hline
\textbf{Feature}     & \textbf{DB1A} & \textbf{DB2A }& \textbf{DB3A} &\textbf{ DB4A} \\ \hline \hline
\(MCC_{SDK}\)   & 1.32 & 1.85 & 14.1 & 4.57  \\ \hline
\(MCC_o\)   & 0.89 & 0.75 & 13.6 & 3.50  \\ \hline
\(MCC_f\) 	& \textbf{0.60} & \textbf{0.53} & 13.1 & \textbf{3.10}   \\ \hline
\(MCC_e\)   & 0.75 & 0.71 & 11.8 & 3.17  \\ \hline
\(MCC_{co}\)  & 0.92 & 0.60 & 13.2 & 3.64  \\ \hline
\(MCC_{cf}\) 	& 0.71 & 0.57& 15.2 & \textbf{3.10}   \\ \hline
\(MCC_{ce}\)  & 0.92 & 0.89 & \textbf{11.6} & 4.10  \\ \hline
\end{tabular}
\end{table}

\begin{table}[h]
 \caption{EERs for FVC 2004 DBs.}
\label{eersFVC2004}
\begin{tabular}{|*{5}{p{1.11cm}|}}
\hline
\textbf{Feature}     & \textbf{DB1A} & \textbf{DB2A }& \textbf{DB3A} &\textbf{ DB4A} \\ \hline \hline
\(MCC_{SDK}\)   & 6.07 & 5.75 & 4.42 & 3.57 \\\hline
\(MCC_o\)   & 5.21 & 6.38 & 6.00 & 3.75  \\ \hline
\(MCC_f\) 	& 4.32 & 6.00 & 6.15 & 2.68   \\ \hline
\(MCC_e\)   & 4.25 & 6.78 & 6.72 & 3.20  \\ \hline
\(MCC_{co}\) &\textbf{3.85} & \textbf{5.35} & \textbf{3.78} & \textbf{2.38}  \\ \hline
\(MCC_{cf}\) & 4.00 & 5.71 & 5.63 & 2.39   \\ \hline
\(MCC_{ce}\)  & 4.24 & 6.46 & 6.74 & 3.07  \\ \hline

\end{tabular}
\end{table}

\begin{table}[h]
\caption{FMR1000 for FVC 2004 DBs.}
\label{1000FMRsFVC2004}
\begin{tabular}{|*{5}{p{1.11cm}|}}
\hline
\textbf{Feature}     & \textbf{DB1A} & \textbf{DB2A }& \textbf{DB3A} &\textbf{ DB4A} \\ \hline \hline
\(MCC_{SDK}\)   & 20.1 & 11.6 & 12.6 & 9.75  \\ \hline
\(MCC_o\)   & 13.2 & 10.4 & 12.4 & 5.71  \\ \hline
\(MCC_f\) 	& \textbf{11.8} & 9.21 & 9.78 & \textbf{5.25 }  \\ \hline
\(MCC_e\)   & 11.9 & \textbf{8.28} & \textbf{8.71} & 5.71  \\ \hline
\(MCC_{co}\)  & 12.9 & 10.9 & 11.92 & 5.92  \\ \hline
\(MCC_{cf}\) 	& 13.1 & 10.03& 11.92 & 5.57   \\ \hline
\(MCC_{ce}\)  & 13.4 & 10.85 & 13.57 & 7.64  \\ \hline
\end{tabular}
\end{table}

The fingerprint matching performance comparison of the different descriptors are presented in the terms of Equal Error Rates (EER) and Detection Error Trade-off (DET) curves. Tables \ref{eersFVC2002}, \ref{eersFVC2004} report the EERs and tables \ref{1000FMRsFVC2002}, \ref{1000FMRsFVC2004} report FMR1000 for all parts [1A-4A] of the FVC 2002 and FVC 2004 dbs. For each part the best EER is highlight in bold. The DET performance curves are reported in figures \ref{DETCurves2002} and \ref{DETCurves2004}.
It can be clearly seen that the experimental results show significant difference in between the variants of the MCC descriptor. Indeed, our variants of MCC, based on texture features, reduce the rate of impostor persons that could be incorrectly accepted (FAR) for most values of FRR in most of cases. Considering the low quality data-sets (FVC 2004) , the DET curves show that MTCC features result in lesser FAR and lesser FRR. Only in the case of FVC 2004 DB3A does \(MCC_o\) actually surpass most of the features and yet lags behind \(MCC_e\) feature.\\
Table \ref{eersFVC2002} shows the EERs for FVC 2002 dbs. MCC-SDK EERs are shown in the top row. It can be clearly seen that the performance of MTCC features are mostly superseding MCC. Even the energy feature is superseding MCC. Recall that we have have the same set of enhanced images, same parameters set and different underlying descriptors that allows us to show the discriminating behavior of texture features. Typically FVC 2002 comprises of fingerprint images that are of a bit better quality, especially DB1A. For the same db, the replacement features are at half the EER of MCC-SDK. Likewise for DB2A, the replacement features are performing even better. For the tougher dbs, DB3A and DB4A, we can see that the EERs are still better than MCC-SDK EERs.

From Table \ref{eersFVC2004}, it is evident that the texture features are out performing not only MCC-SDK features, but there is also intra-competition between the texture features with \(MCC_{co}\) being the highest performer in terms of lowest EER. This shows that orientation is a better feature when the database comprises of fingerprint images that are of low quality just.. 
\begin{figure*}[h]
\centering
\subfloat[ROC curves for FVC 2002 DB1A ]
{\label{det2002db1a}\includegraphics[height=2in]{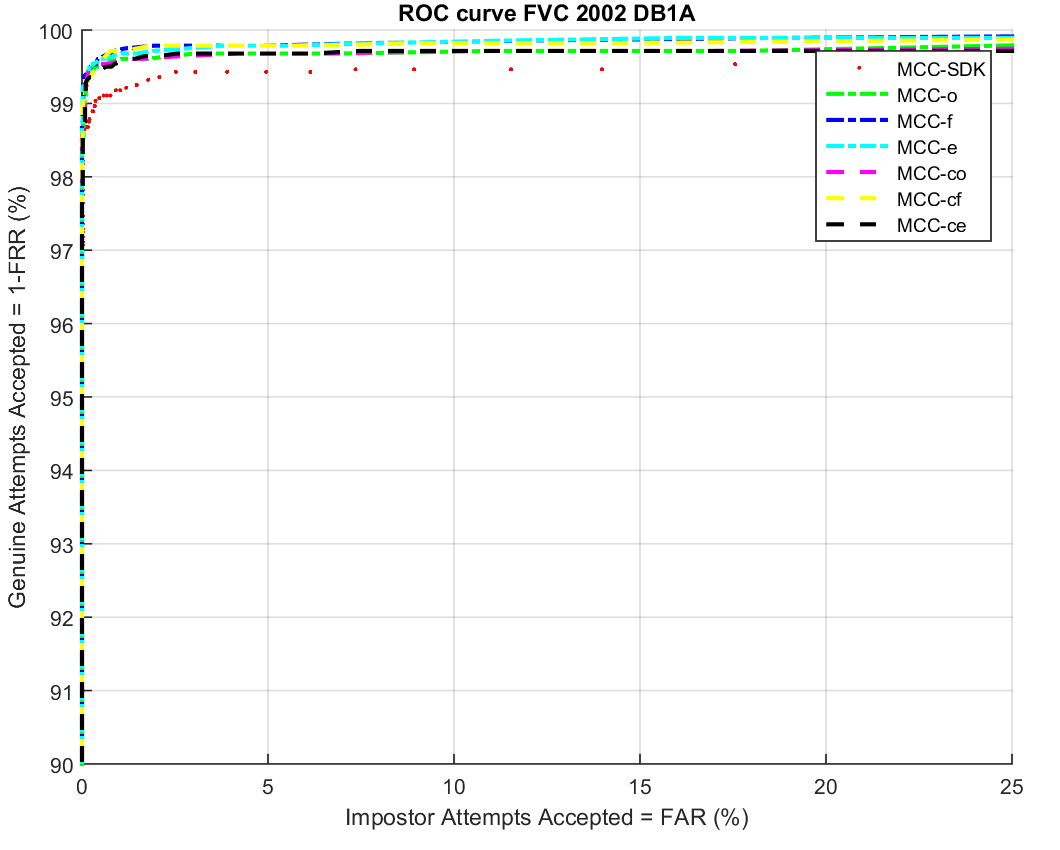}}\thinspace
\subfloat[ROC curves for FVC 2002 DB2A ]{\label{det2002db2a}\includegraphics[height=2in]{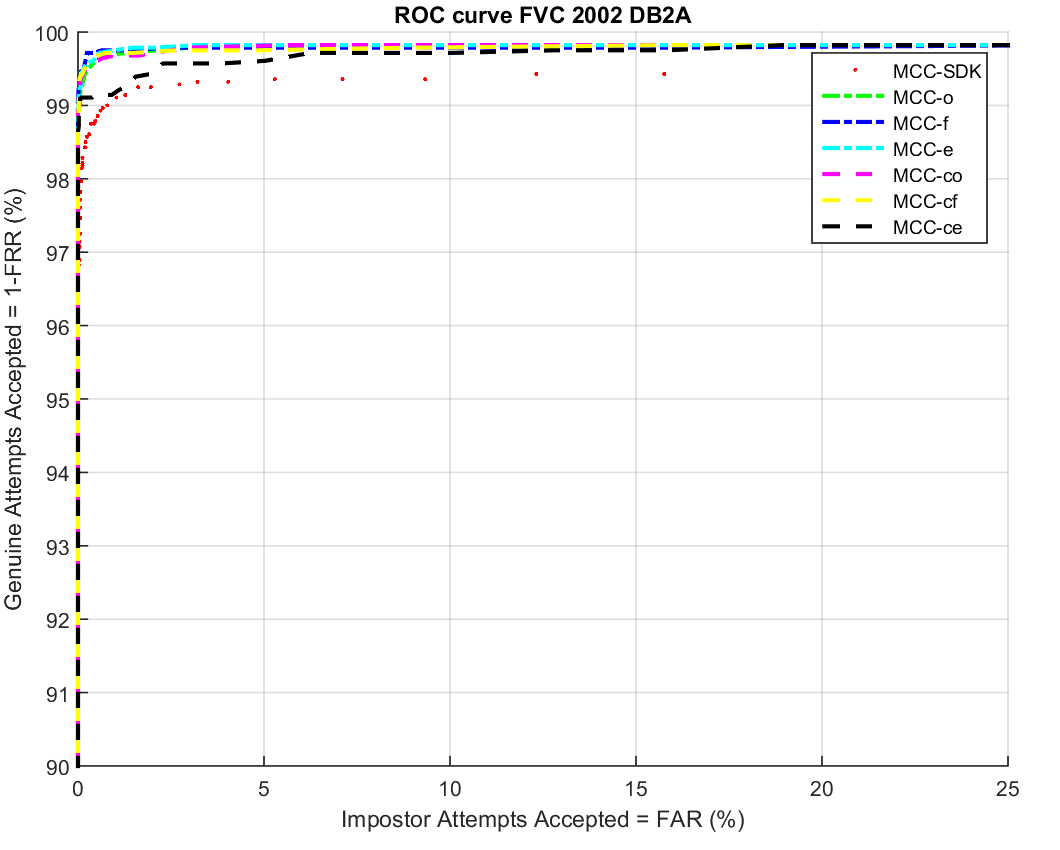}}
\\
\subfloat[ROC curves for FVC 2002 DB3A ]
{\label{det2002db3a}\includegraphics[height=2in]{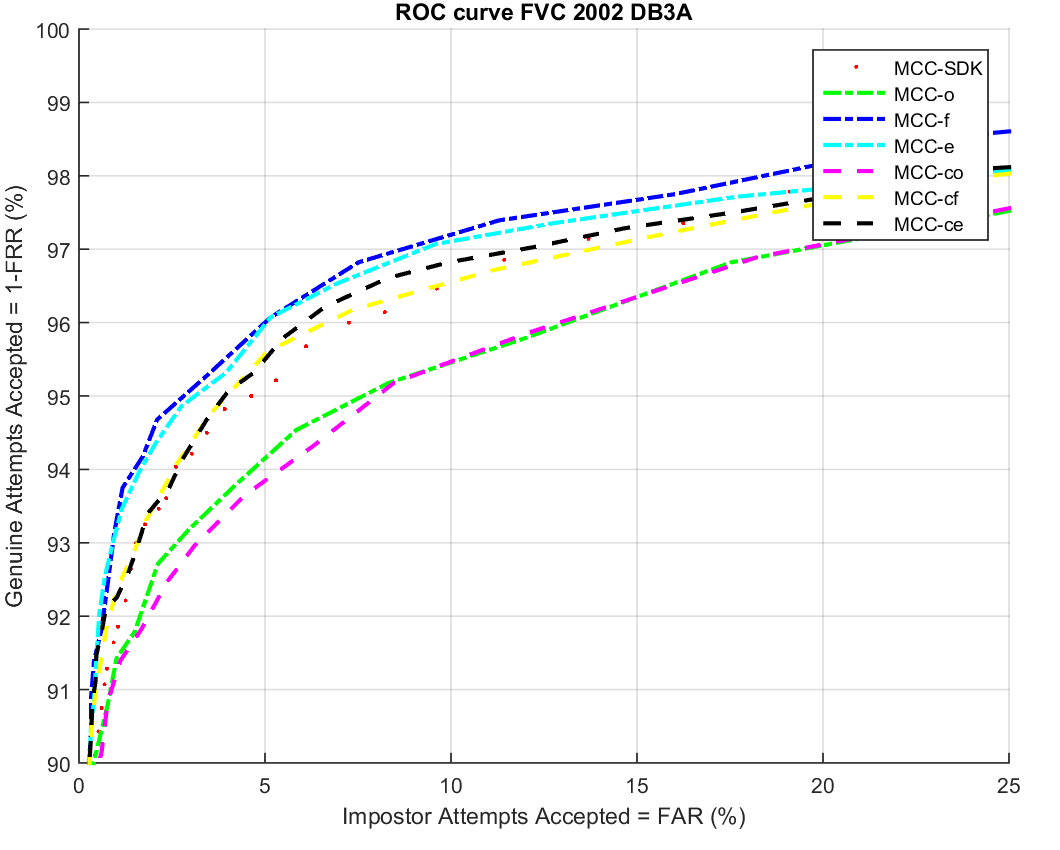}}\thinspace
\subfloat[ROC curves for FVC 2002 DB4A ]{\label{det2002db4a}\includegraphics[height=2in]{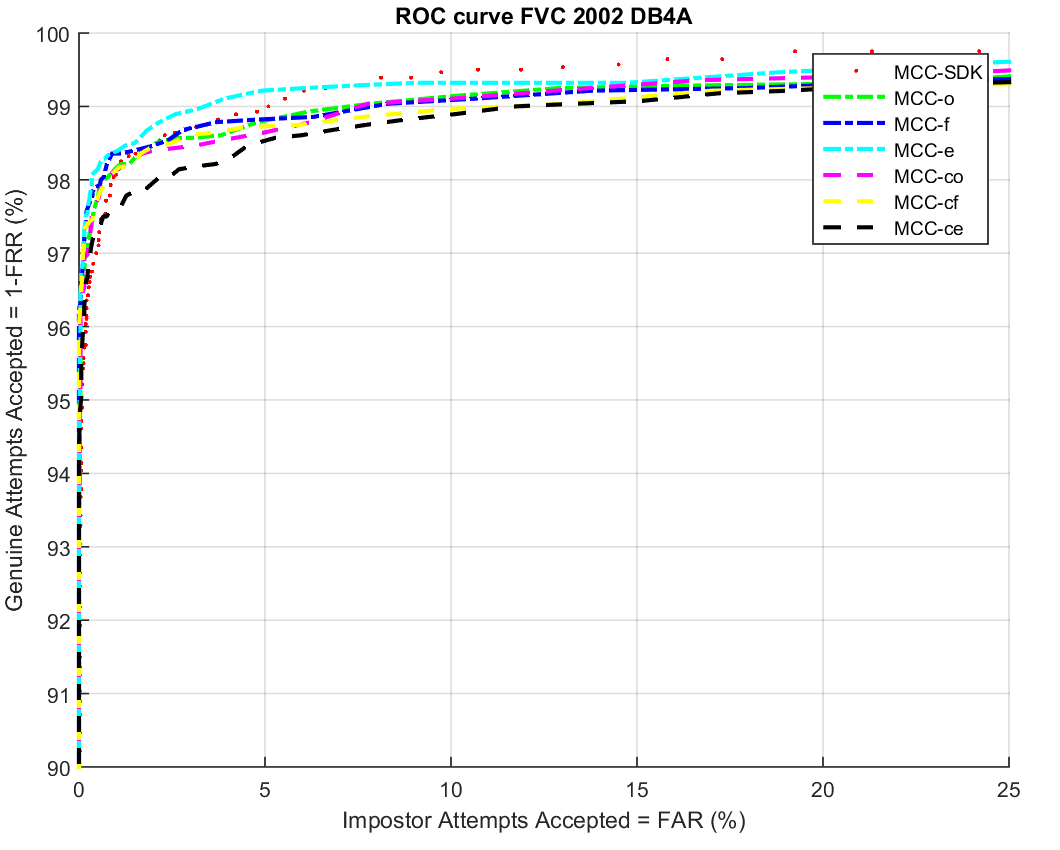}}
\\
\caption{{\bf ROC curves for 2002 DBs} The (x,y)  axes are limited for clarity.}

\label{ROC2002}
\end{figure*}

\begin{figure*}[h]
\centering
\subfloat[ROC curves for FVC 2004 DB1A ]
{\label{roc2004db1a}\includegraphics[height=2in]{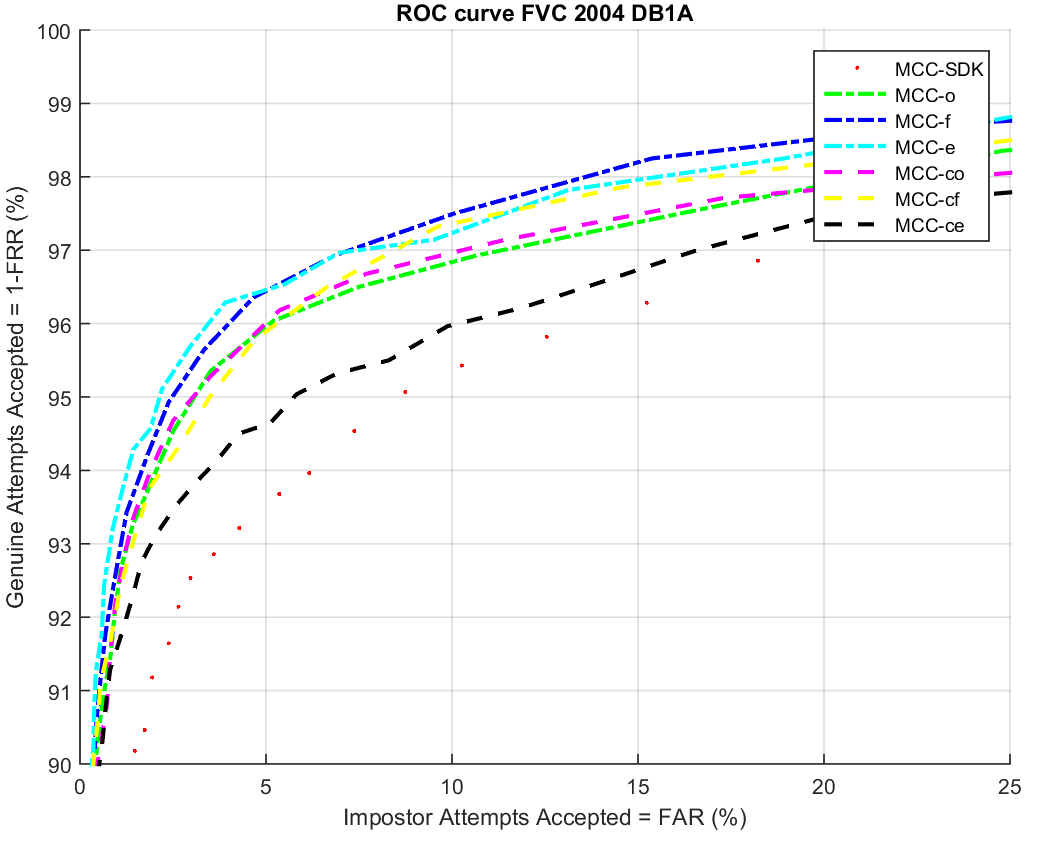}}\thinspace
\subfloat[ROC curves for FVC 2004 DB2A ]{\label{roc2004db2a}\includegraphics[height=2in]{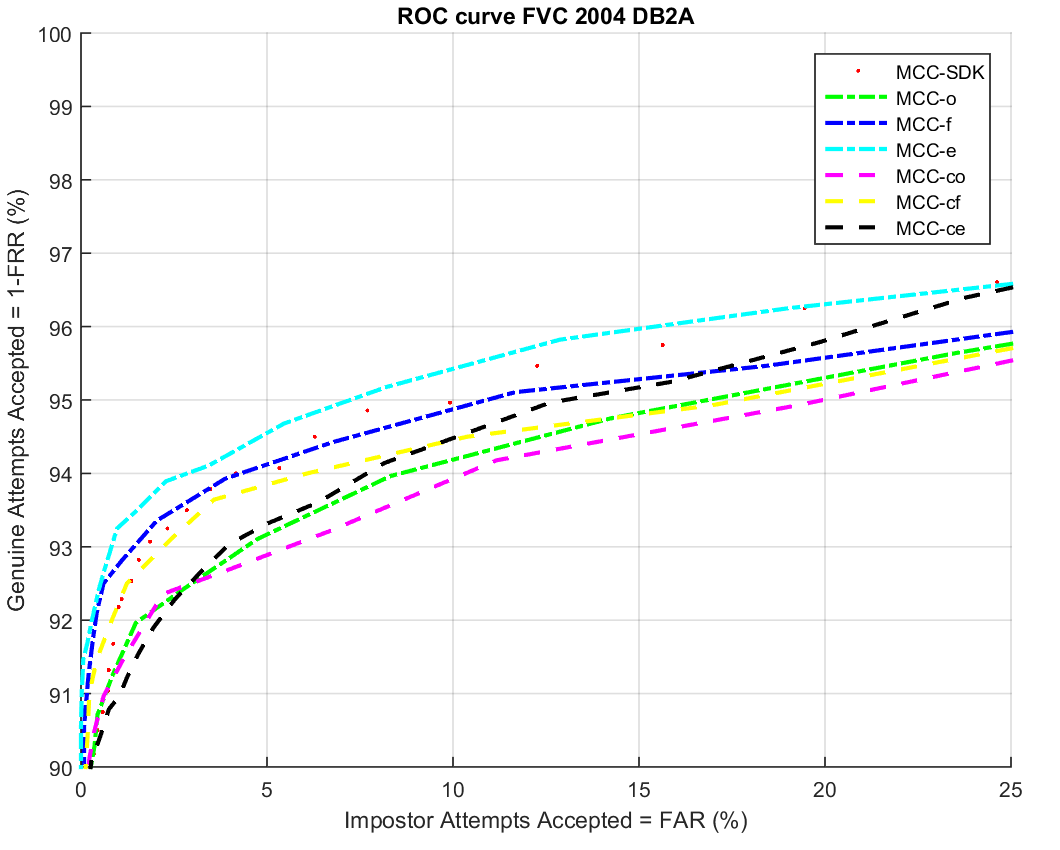}}
\\
\subfloat[ROC curves for FVC 2004 DB3A ]
{\label{roc2004db3a}\includegraphics[height=2in]{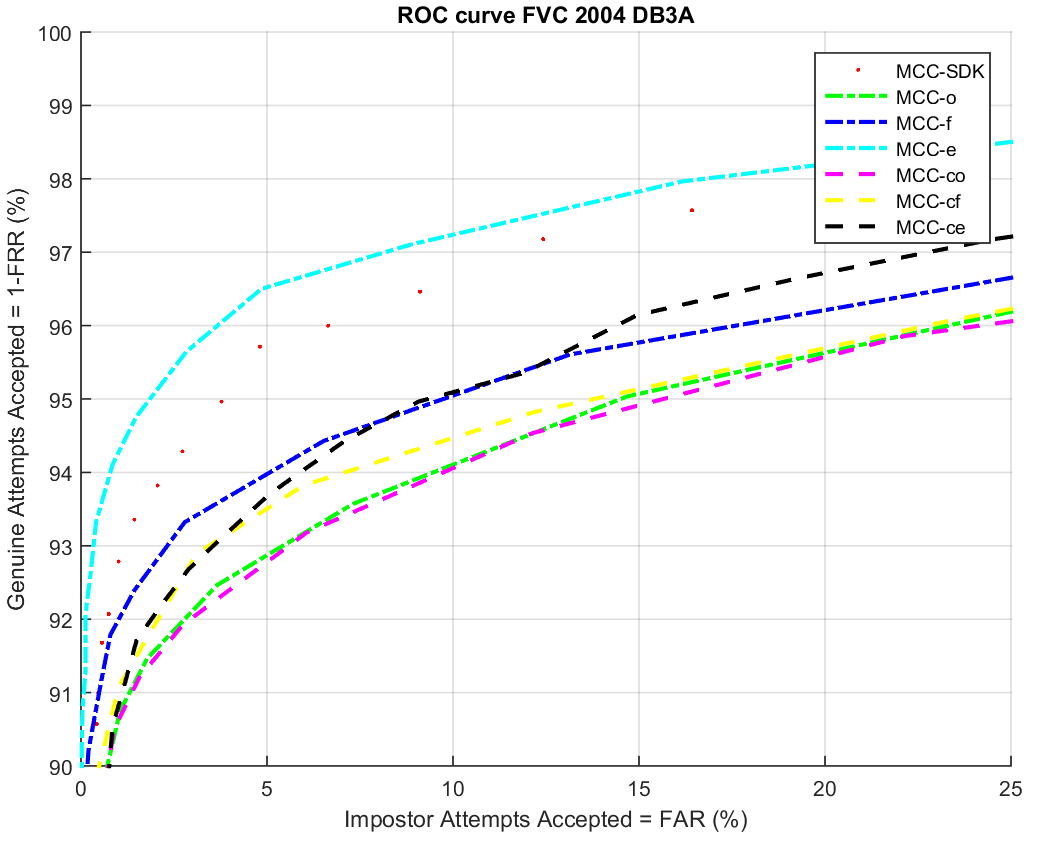}}\thinspace
\subfloat[ROC curves for FVC 2004 DB4A ]{\label{roc2004db4a}\includegraphics[height=2in]{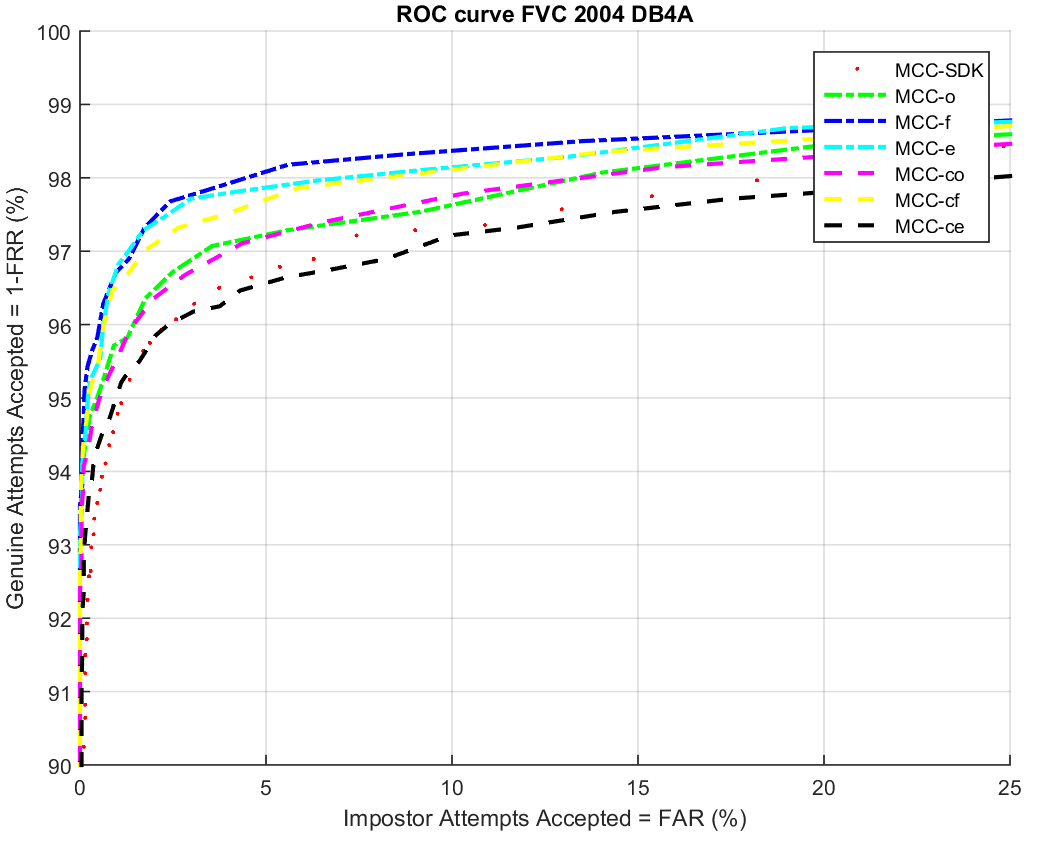}}
\\
\caption{{\bf ROC curves for 2004 DBs} The (x,y)  axes are limited for clarity.}

\label{ROC2004}
\end{figure*}
This performance should not be surprising. The cell-centered features are performing better as each cell is picking up the valuable information around the central minutia from the texture images (orientation, frequency and energy). This works like a sampling method at fixed intervals. The cell centers being located at fixed intervals from each other are able to sample the underlying texture information.  This aspect of sampling is not a part of the original MCC or the texture features \(MCC_o\),\(MCC_f\) and \(MCC_e\). 
This means that the cells in the MCC descriptor will now have additional discriminatory information when matching is performed.
\begin{figure}[h]
\centering
\subfloat[Subject 75, sample 4 from FVC 2002 DB1A]
{\label{det2002db1a}\includegraphics[height=1.6in]{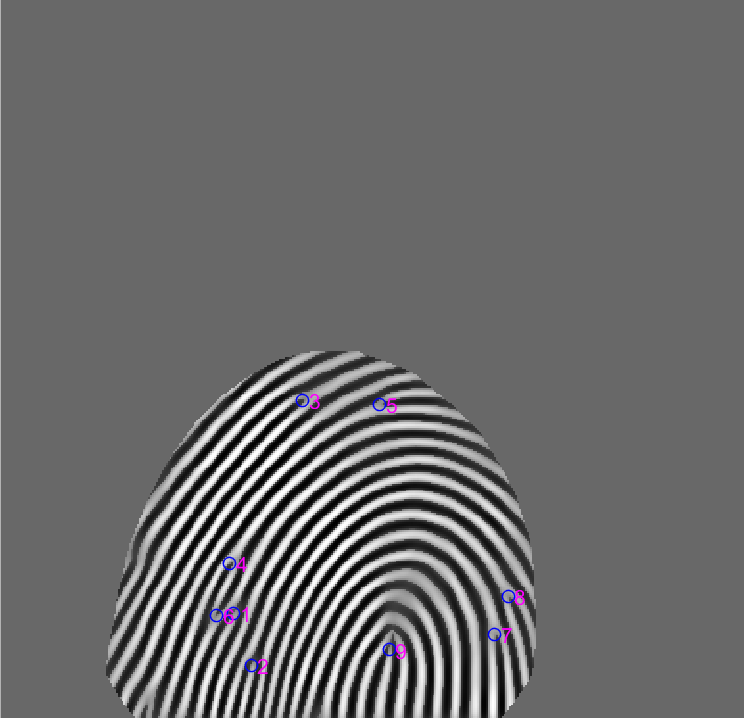}}\thinspace
\subfloat[Subject 75, sample 7 from FVC 2002 DB1A ]{\label{det2002db2a}\includegraphics[height=1.6in]{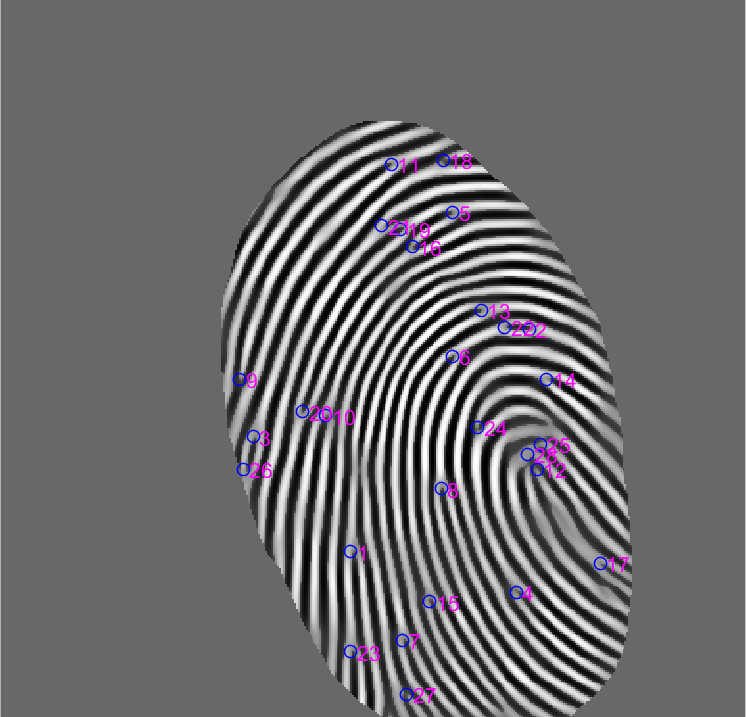}}

\caption{{\bf Two samples from FVC 2002 DB1A} Minutiae for a partial image and a rotated full image. }

\label{GenuinePairs}
\end{figure}
To elaborate the sampling affect, the local and global matching scores can be understood. Figure \ref{GenuinePairs} shows two samples from FVC 2002DB1A with all their minutiae. After performing a local similarity sort, we are left with high scoring minutiae pairs which are then passed onto relaxation. These pre-relaxation scores are listed in Table \ref{matchpairsFVC2002} for all the features.

\begin{table}[h]
 \caption{Matched/Paired minutiae for 75\_4 and 75\_7 of FVC 2002 DB1A.}
\label{matchpairsFVC2002}
\begin{tabular}{c|*{7}{p{0.85cm}}}
\hline
\textbf{ Pair}     & \textbf{\(MCC_o\)} & \textbf{\(MCC_f\)}& \textbf{\(MCC_e\)} &\textbf{\(MCC_{co}\)} & \textbf{\(MCC_{cf}\)} & \textbf{\(MCC_{ce}\)}\\ \hline \hline
2,1   & 0.41 & 0.47 & 0.48 & 0.48 & 0.48 & 0.48 \\\hline
9,18  & 0.26 & 0.41 & 0.41 & 0.41 & 0.41 & 0.41 \\\hline
9,19  & 0.15 & 0.42 & 0.43 & 0.42 & 0.42 & 0.43 \\\hline
9,20  & 0.24 & 0.47 & 0.47 & 0.46 & 0.47 & 0.47 \\\hline

\end{tabular}
\end{table}

\begin{table}[h]
 \caption{Matching scores for samples from FVC 2002 DB1A. Final global matching scores using LSSR.}
\label{matchSamplesFVC2002}
\begin{tabular}{c|*{7}{p{0.75cm}}}
\hline
\textbf{Case}     & \textbf{\(MCC_o\)} & \textbf{\(MCC_f\)}& \textbf{\(MCC_e\)} &\textbf{\(MCC_{co}\)} & \textbf{\(MCC_{cf}\)} & \textbf{\(MCC_{ce}\)}\\ \hline \hline
75\_5 , 75\_7   & 0.053 & 0.078 & 0.080 & 0.069 & 0.077 & 0.080 \\\hline
1\_1 , 12\_1  & 0.070 & 0.067 & 0.062 & 0.067 & 0.065 & 0.065 \\\hline
\end{tabular}
\end{table}

Since the feature \(MCC_o\) is our implementation of MCC, we can make a in-depth comparison with MTCC features.
As the reported scores suggest (Table \ref{matchpairsFVC2002}), the MTCC features are performing better in case of the minutiae pairs \{9,18\},\{9,19\} and \{9,20\}. These pairs are visually confirmed to be structurally correct pairs. This means that the texture features \(MCC_f\) and \(MCC_e\) are more discriminant than \(MCC_o\) and additional texture information in cell centered features are helping in propping up the genuinely matching minutiae scores. This particular case relates to partial fingerprint matching where one fingerprint template \((75\_4\)) is a partial acquisition.

Table \ref{matchSamplesFVC2002} shows two global matching scores.  One a genuine match score and the other an impostor match score. 
The table reports a higher matching score for all texture features compared to \(MCC_o\). Exhibiting such high scores in case of partial matching is very important as partial match cases generally tend to produce lower scores. The examples of matching 75\_4 with 75\_7 is also a partial fingerprint matching case. For the impostor match, the scores from texture features are minutely lower. This affect is still less as compared to the genuine score improvements. This means that genuine scores are pushed to a higher level while impostor scores are affected a little less.

The final performance indicators of ROC are reported in Figures \ref{ROC2002} and \ref{ROC2004}. FVC 2002 data-set is comparatively a better quality DB as comparted to FVC 2004. The ROC curves indicate closely related performance except in the case of FVC 2002 DB3A indicative of a DB with low quality.
Now considering the case of energy and frequency features in FVC 2004 DB3A, energy feature has the highest indicator of performance. Likewise frequency is the highest indicator of performance in FVC 2004 DB4A. These performance cues show the improved performance of MCC when used in terms of MTCC features. 

\section{Conclusion}
\label{conclusion}
Improving the performance of fingerprint matching is an active domain of research. The main focus of any research would be improving descriptors and matching methods. Keeping the descriptor and matching methods same, we provided a novel change to the underlying content of the well know MCC descriptor to improve the aggregated fingerprint matching accuracy. From a minutiae only descriptor to a texture descriptor with higher discriminatory properties was presented. \\
Out of the three different texture discriminant, orientation features are more often the best performers make a strong relation to minutia angular information. 
Experimental results show that the cell-centered features \(MCC_{co}\) shows highest performance on FVC 2004 database which retains low quality fingerprints.  For the more cleaner data-sets FVC 2002 DB1 - features \(MCC_o\) and \(MCC_f\) outperform traditional MCC. 
To our knowledge, there is no literature that shows the underlying content of MCC being modified to such an extent that it can outperform original MCC features. The use of energy as a feature is not new but in the aspect of MCC, this is unique. 

As part of future prospects,We intend to work on a version that combines descriptors at feature level or score level. Since there are six different descriptor, each descriptor can be combined with the other to achieve a better performance.  Combining with MCC traditional features is also an option that can be useful and is to be studied. A sped up version of the proposed features is also a consideration.

\section*{Acknowledgment}
This research has been supported by the Technology \& Development Directorate of National Database And Registration Authority (NADRA).

\ifCLASSOPTIONcaptionsoff
  \newpage
\fi



%

%
\vspace{-1cm}
\begin{IEEEbiography}
[{\includegraphics[width=1in,height=1.25in,clip,keepaspectratio]{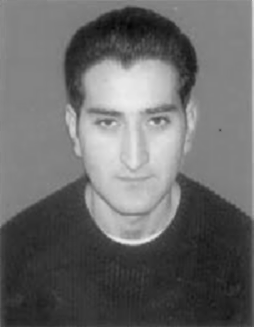}}]{Wajih Ullah Baig}
Wajih Ullah Baig received his B.S Software Development (CS) degree from Hamdard University in 2005. He has a decade of experience in RnD and software development. He has worked on numerous projects related to computer vision and image processing inclusive of CBIR, Biometrics, Tracking and Video Analytics. His passion and interests are in research in Computer Vision, Deep Learning, Machine Learning, Algorithms.
\end{IEEEbiography}
\vspace{-1cm}
\begin{IEEEbiography}
[{\includegraphics[width=1in,height=1.25in,clip,keepaspectratio]{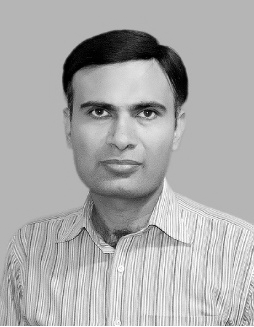}}]{Umar Munir}
Dr. Muhammad Umar Munir received his B.E. Telecommunications Engineering form National University of Sciences and Technology (NUST), Pakistan in 1998. He obtained his M.S and Ph.D in Computer Software Engineering from NUST in 2004 and 2012 respectively. He is a member of IEEE, USA and IEEE Computer Society, USA. He has over 10 years of research and development experience in Automatic Fingerprint Identification Systems.  His research interest includes Biometrics and Computer Vision.
\end{IEEEbiography}
\vspace{-1cm}
\begin{IEEEbiography}
[{\includegraphics[width=1in,height=1.5in,clip,keepaspectratio]{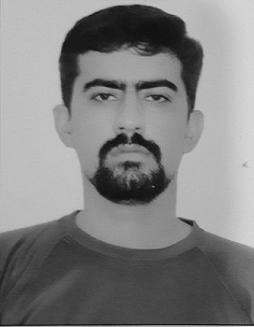}}]{Kashif Sardar}
Kashif Sardar received his B.S.c Software Engineering degree from University of engineering and technology Taxila in 2012.Recentlty he has completed M.S. of Software Engineering from University of engineering and technology Taxila. He has 5 years’ of experience in software development. He has worked on distributed fingerprint recognition system. His research interests are in Distributed system and Machine Learning\end{IEEEbiography}
\vspace{-1cm}
\begin{IEEEbiography}
[{\includegraphics[width=1in,height=1.25in,clip,keepaspectratio]{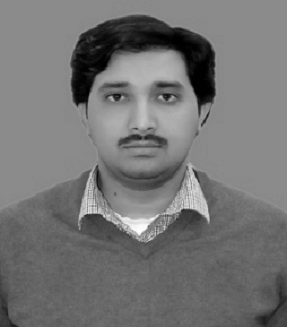}}]{Waqas Ellahi}
Waqas Ellahi received the bachelor’s degree in Electronic Engineering from the Mohammad Ali Jinnah University (MAJU) in 2010, the master’s in Electrical Engineering from Center For Advanced Studies in Engineering (CASE) in 2012. He has more than 6 years of industrial experience in Biometric, Video Codec and Mobile application Development. His research interests include Computer Vision, pattern recognition and Deep Learning algorithms.\end{IEEEbiography}
\vspace{-1cm}
\begin{IEEEbiography}
[{\includegraphics[width=1in,height=1.25in,clip,keepaspectratio]{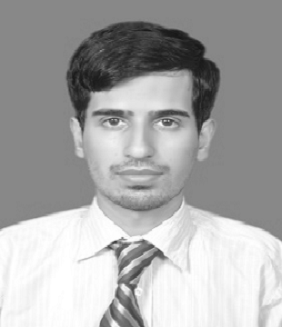}}]{Adeel Ejaz}
Adeel Ejaz received his B.E. Computer Engineering degree from Bahria University Islamabad campus in 2013.Recentlty he has completed M.S. of Computer Engineering from National University of Sciences and Technology (NUST). He has 2 years’ experience in software development. He has worked on Pattern Recognition and image processing related Projects. His research interests are in Image Processing, Pattern Recognition, Computer Vision and Machine Learning\end{IEEEbiography}



\end{document}